\documentclass{article}

\usepackage[utf8]{inputenc}
\usepackage{textcomp}
\DeclareUnicodeCharacter{266B}{\textmusicalnote}

\usepackage{microtype}
\usepackage{graphicx}
\usepackage{subcaption}
\usepackage{booktabs} % for professional tables

\usepackage{hyperref}

\usepackage{enumitem}

\usepackage[preprint]{icml2026}

\usepackage{amsmath}
\usepackage{amssymb}
\usepackage{mathtools}
\usepackage{amsthm}

\usepackage[capitalize,noabbrev]{cleveref}

\theoremstyle{plain}

\theoremstyle{definition}

\theoremstyle{remark}

\usepackage{multirow}

\usepackage[textsize=tiny]{todonotes}

\newcommand{\MyModel}{ScalDPP}

\begin{document}

\twocolumn[
  % \icmltitle{DPP-Driven Reranking: Enhancing Multi-Hop Retrieval-Augmented Generation Beyond Semantic Similarity}

  % \icmltitle{From Similarity to Diversity: DPP-Powered Rerankers for Multi-Hop RAG}
  
  % \icmltitle{From Redundant Chunks to Complementary Context: Scalable DPPs for RAG}
  \icmltitle{Scaling DPPs for RAG: Density Meets Diversity}
%Enhancing Retrieval-Augmented Generation with Scalable DPPs: Modeling Inter-Chunk Diversity and Complementarity
%From Redundant Chunks to Complementary Context: Scalable DPPs for RAG
% author list: Xu Sun, Banghen Xie, Qiang Gao, Li Huang



  \icmlsetsymbol{equal}{*}

  \begin{icmlauthorlist}
    \icmlauthor{Xun Sun}{sch,comp}
    \icmlauthor{Baiheng Xie}{sch,comp}
    \icmlauthor{Li Huang}{sch}
    \icmlauthor{Qiang Gao}{sch}
  \end{icmlauthorlist}

  \icmlaffiliation{sch}{Southwestern University of Finance and Economics, Chengdu, China}
  \icmlaffiliation{comp}{Zhida AI, Zhida Technology, Chengdu, China}

  \icmlcorrespondingauthor{Li Huang}{lihuang@swufe.edu.cn}

  \icmlkeywords{Machine Learning, ICML}

  \vskip 0.3in
]



\printAffiliationsAndNotice{}  % no special notice (required even if empty)

\begin{abstract}
Retrieval-Augmented Generation (RAG) enhances Large Language Models (LLMs) by grounding generation in external knowledge, yielding relevance responses that are aligned with factual evidence and evolving corpora. Standard RAG pipelines construct context through relevance ranking, performing point-wise scoring between the user query and each corpora chunk. This formulation, however, ignores interactions among retrieved candidates, leading to redundant contexts that dilute density and fail to surface complementary evidence. We argue that effective retrieval should optimize jointly for both density and diversity, ensuring the grounding evidence that is \textit{dense in information yet diverse in coverage}. In this study, we propose~\MyModel, a diversity-aware retrieval mechanism for RAG that incorporates Determinantal Point Processes (DPPs) through a lightweight P-Adapter, enabling scalable modeling of inter-chunk dependencies and complementary context selection. In addition, we develop a novel set-level objective, Diverse Margin Loss (DML), that enforces ground-truth complementary evidence chains to dominate any equally sized redundant alternatives under DPP geometry. Experimental results demonstrate the superiority of \MyModel, substantiating our core statement in practice.

\end{abstract}

\section{Introduction}
\label{intro}
\begin{figure}[t]
\centering
\includegraphics[width=\linewidth]{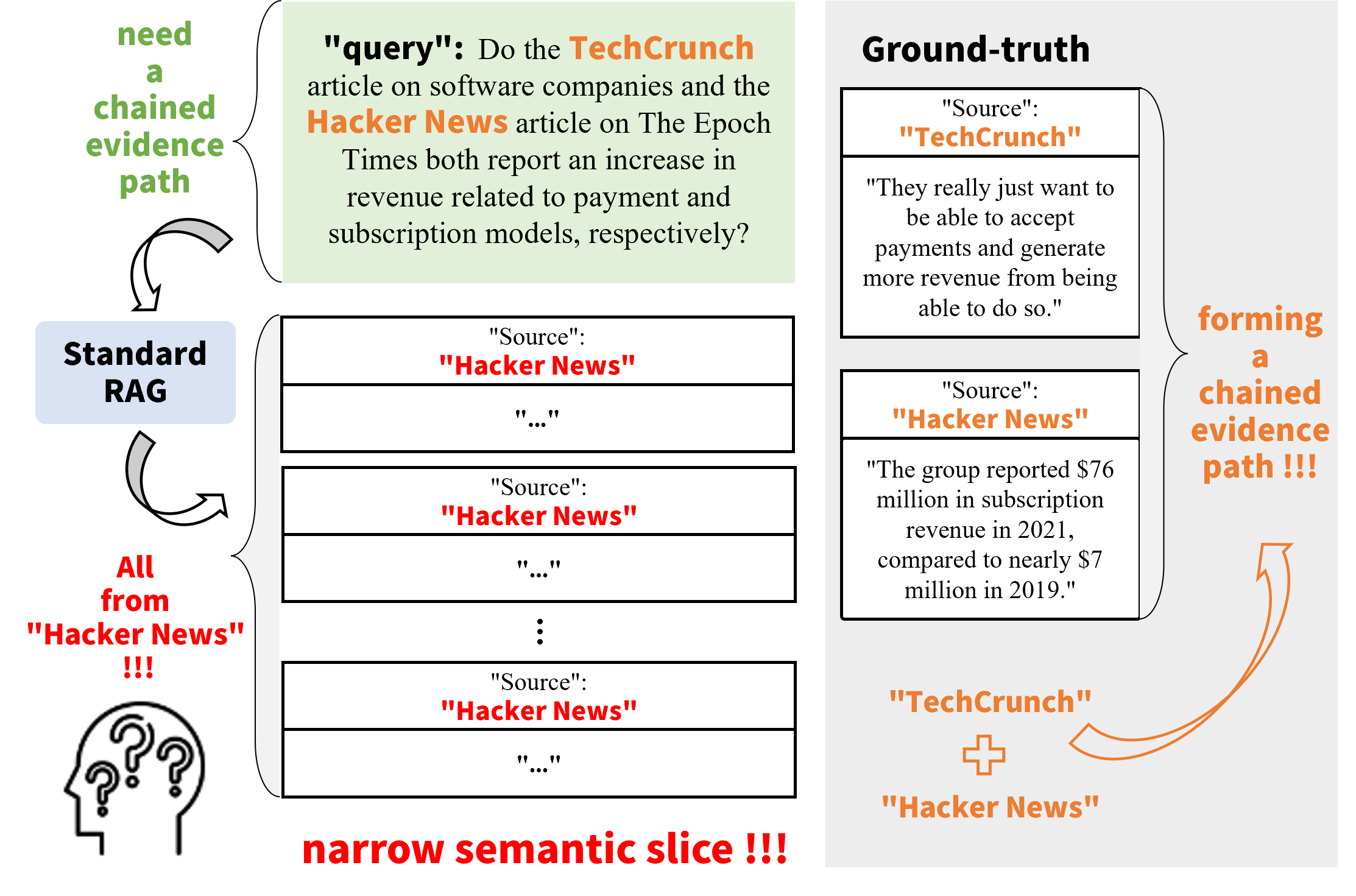}
\caption{Standard RAG models query-chunk relevance but neglects inter-chunk diversity and complementarity. Example from MultiHop-RAG~\cite{tang2024multihoprag}.}
\label{fig:motivation}
\vspace{-0.5cm}
\end{figure}

Large language models (LLMs) have achieved strong performance across a wide range of natural language understanding and generation tasks. Nevertheless, they remain fundamentally constrained by the probabilistic autoregressive nature, which prioritizes textual coherence over factual accuracy, 
leading to inappropriate outputs, such as hallucinated or outdated content~\cite{trivedi2023interleaving}. Retrieval-Augmented Generation (RAG) mitigates these limitations by dynamically retrieving and incorporating external, domain-specific knowledge during the generation process~\cite{lewis2020retrieval, guu2020realm}, enabling relevance-aware responses that are better grounded in factual information. 

% In standard RAG pipelines, this is typically implemented by partitioning the corpus into fixed-length textual chunks and retrieving a small set of relevant candidates via embedding-based similarity matching between the input query and the chunks, which are then concatenated and provided to the LLM as additional context~\cite{gao2024retrievalaugmentedgenerationlargelanguage,fan2024survey}. Optionally, a reranker assigns refined relevance scores and reorders candidates. While in practice, they implicitly assume that the highest-scoring chunks are both sufficient and mutually informative. The efficacy of RAG systems is critically dependent on the \texttt{informational density and uniqueness} of retrieved context. %Redundancy is not an edge case, but a natural consequence of relevance-based retrieval. 
In standard RAG pipelines, the corpus is partitioned into fixed-length textual chunks, from which a small set of relevant candidates is retrieved via embedding-based similarity matching with the input query~\cite{gao2024retrievalaugmentedgenerationlargelanguage,fan2024survey}. These candidates are then concatenated and provided to LLMs as additional context, optionally followed by a reranking stage that refines relevance scores. Such pipelines implicitly assume that the top-ranked chunks are sufficient. However, these candidates are selected based on their similarity to the input query, which inevitably produces clusters of near-duplicate chunks, such as multiple paraphrases of the same fact. Under a limited context window, such redundancy dilutes the effective token budget and constrains the generator to reason over a narrow semantic slice~\cite{hsieh2024ruler, wang-etal-2024-retrieve, wang2025retrievalaugmentedquestionanswering}.
Moreover, similarity-driven retrieval further overlooks chunks that are individually weaker matches but collectively essential, as it fails to account for orthogonal attributes, latent constraints, or cross-cutting perspectives required for multi-hop reasoning. Thus, what appears as evidence can be misleading: redundant chunks crowd out uniquely informative context. As a result, candidates are highly correlated with the query, while inter-candidate interactions -- particularly diversity and complementarity -- are insufficiently captured, as illustrated in Fig.~\ref{fig:motivation}. Although recent approaches leverage knowledge graphs to model entity-level interactions through structured relational paths~\cite{edge2025localglobalgraphrag, guo2025lightragsimplefastretrievalaugmented}, they typically depend on costly graph pre-construction.
Furthermore, they emphasize explicit entity links rather than probabilistic optimization over chunk-level subsets, limiting scalability and flexibility in RAG settings~\cite{li2026pankragenhancinggraphretrieval}.
Motivated by these analyses, we ask whether retrieval in RAG can be reformulated 
% in a more principled manner 
by explicitly ensuring that the grounding evidence is \textit{dense in information yet diverse in coverage}.
% accounting for the need for informative and non-redundant context.
To this end, we are the first to employ Determinantal Point Processes (DPPs)~\cite{macchi1975coincidence, hough2006determinantal, kulesza2012learningdeterminantalpointprocesses}, a class of probabilistic models rooted in statistical physics and random matrix theory, into RAG systems. DPPs naturally model subset-level diversity through negative correlations, providing a principled foundation for constructing informative and complementary contexts beyond relevance-based retrieval.
% Motivated by the need for dense yet unique context, we propose~\textbf{\MyModel}, which incorporates \textbf{Scal}able \textbf{D}eterminantal \textbf{P}oint \textbf{P}rocesses (DPPs)~\cite{macchi1975coincidence, hough2006determinantal, kulesza2012learningdeterminantalpointprocesses} into RAG systems. In detail, typical DPPs are probabilistic subset selection models that encourage diversity via implicitly modeling negative correlations, which could further XXX.
However, directly applying DPPs to RAG poses two major challenges. First, pre-training the kernel matrix $\mathbf{L}$ in DPPs is computationally prohibitive, incurring $\mathcal{O}(|\mathcal{D}|^2)$ storage %and high time complexity for operations, 
where $\mathcal{D}$ refers to %is the total number of chunks in the 
knowledge base, thus severely restricting scalability when the knowledge base evolves through incremental updates. Moreover, since $\mathbf{L}$ is constrained to be symmetric and positive semi-definite, DPPs can only induce negative dependencies among chunks: increasing similarity between 
% the modeled correlation between any 
$\mathbf{D}_i$ and $\mathbf{D}_j$ necessarily reduces $\det (\mathbf{L}_{\mathbf{D}_i,\mathbf{D}_j})$. Consequently, DPPs are limited to capturing repulsive interactions and cannot express attractive relations that are often essential.
% is constrained to $-L_{ij} L_{ji} = -(L_{ij})^2 \leq 0$. This inherently limits DPPs to capturing only repulsive (diversity-promoting) interactions, while precluding attractive (complementarity-enhancing) relations that could be beneficial in the candidate pool. 

% To overcome both challenges, we propose~\textbf{\MyModel}, a \textbf{Scal}able \textbf{D}eterminantal \textbf{P}oint \textbf{P}rocesses (DPPs) into RAG systems. We replace global pre-training of $L$ by attaching to the base embedding model a parameter-efficient adapter that serves as a lightweight mapping function. During initial retrieval, the adapter remains disabled to preserve the original query-chunk relevance; it is activated only during the subset selection phase, where it injects learned inter-chunk interaction patterns, such as complementarity, into the embeddings. At inference time, we dynamically construct the kernel $L$ over the candidate pool, optionally fusing it with a quality matrix $Q$ derived from reranker scores (defaulting to the identity matrix when reranking is disabled). Subset selection is then performed via Maximum a Posteriori (MAP) inference to obtain a diverse and complementary context.
To overcome both limitations, we propose~\textbf{\MyModel}, a diversity-aware retrieval mechanism for RAG under DPPs geometry.
% a \textbf{Scal}able \textbf{D}eterminantal \textbf{P}oint \textbf{P}rocess framework that can be seamlessly integrated into existing RAG pipelines. 
%Instead of globally pre-training the kernel matrix $\mathbf{L}$ in DPPs, 
\MyModel~attaches a parameter-efficient P-Adapter to the base embedding model, serving as a lightweight mapping function. During initial retrieval, P-Adapter is disabled to preserve the original query–chunk relevance, and is activated only during subset selection to inject learned inter-chunk interaction patterns %, such as complementarity, 
into the embeddings. When inference, \MyModel~dynamically constructs the kernel $\mathbf{L}$ over the retrieved candidate pool, optionally fusing it with a quality matrix $\mathbf{Q}$ derived from reranker scores (otherwise $\mathbf{Q}=\mathbf{I}$).
% defaulting to the identity matrix when reranking is absent).
Subset selection is then performed via Maximum a Posteriori (MAP) inference, yielding a context that is both diverse and complementary. In addition, we develop a novel set-level objective, Diverse Margin Loss (DML), to optimize the P-Adapter and shape the embedding space such that determinant maximization corresponds to selecting dense yet complementary contexts. 
% Extensive experiments on the MultiHop-RAG benchmark show the superiority of \MyModel. % which enforces ground-truth complementary evidence chains to dominate any equally sized redundant alternatives under DPP geometry. 

Our main contributions are summarized as: 1) We introduce~\MyModel, the first plug-and-play module to extend DPP-based modeling to RAG, explicitly capturing inter-chunk diversity and complementarity beyond query–chunk relevance, 2) Unlike classical DPP formulations, we propose a scalable dynamic kernel construction mechanism coupled with an adaptive embedding adapter P-Adapter to overcome DPPs’ inherent scalability and correlation limitations, enabling complementarity-aware chunk selection, and 3) In contrast to employing standard negative log-likelihood loss (NLL), we develop a novel Diverse Margin Loss (DML) to optimize the proposed P-Adapter, with smooth surrogate formulations that ensure differentiability and favorable optimization properties.

\section{Method}
\label{method}
% In this section, we detail the proposed~\MyModel~approach, which serves as a plug-and-play enhancement module for integrating Determinantal Point Processes (DPPs) into Retrieval-Augmented Generation (RAG) to explicitly model inter-chunk diversity and complementarity. 
% Building on the preliminaries in Section~\ref{preliminary}, we first provide an overview of the approach, followed by specifics on DPP-based subset selection, the adaptive embedding adapter, and the Diverse Margin Loss (DML) for optimization.

% Given a user query $\mathbf{q}$ and a knowledge base $\mathcal{D} = \{\mathbf{D}_1, \dots, \mathbf{D}_m\}$ consisting $m$ textual chunks, embedding models such as BGE-M3~\cite{chen2024bge}) map both $\mathbf{q}$ and $\mathcal{D}$ into a unify semantic space. %$\mathbb{R}^d$. 
% The retrieval process then produces a candidate set $\mathcal{D}^c \subseteq \mathcal{D}$ %\textcolor{purple}{($|C|=N$)} 
% % via efficient approximate nearest neighbor search (e.g., HNSW~\cite{malkov2018efficient}), which is used to augment generators with relevant external knowledge for response.  

\begin{figure*}[t]
    \centering
    \includegraphics[width=\linewidth]{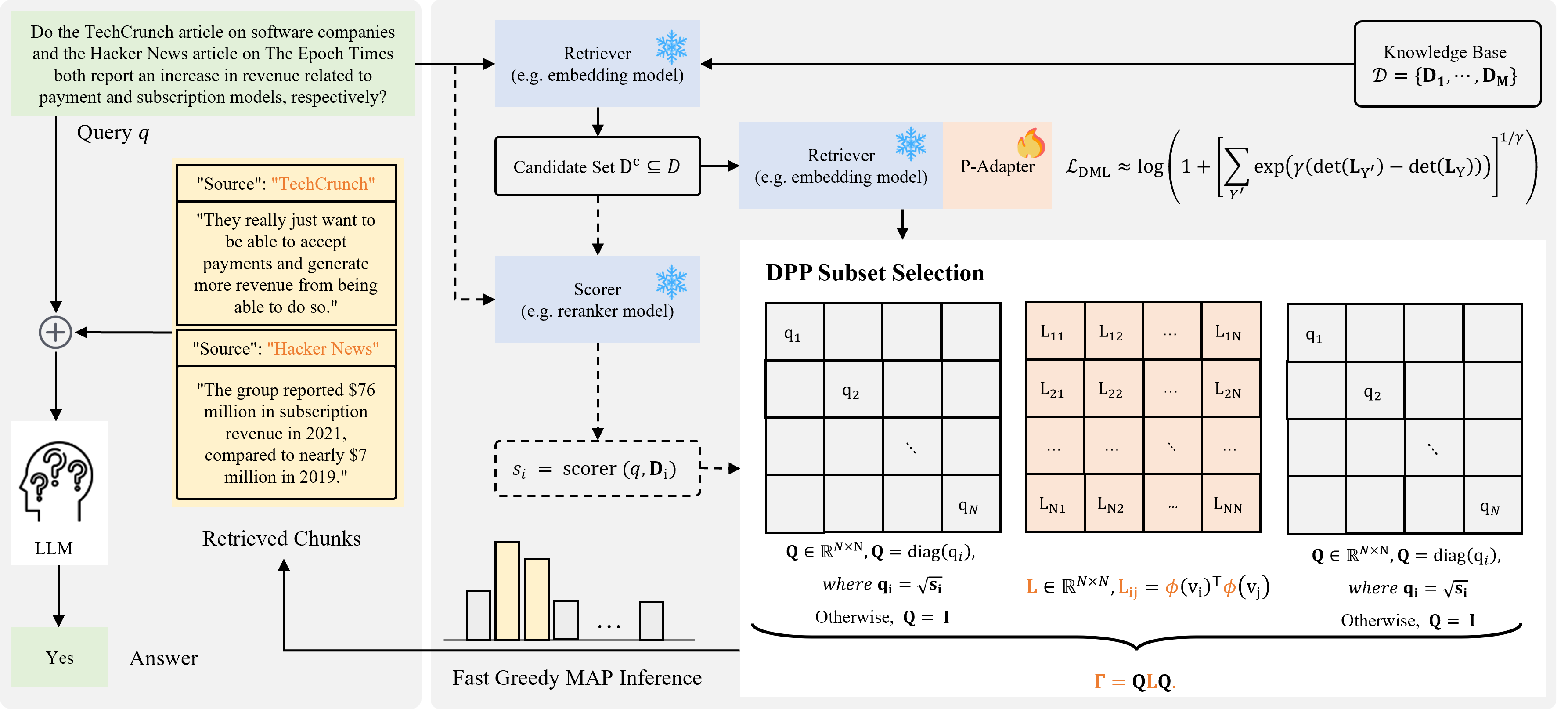}
    \caption{Overview of the~\MyModel~approach. The pipeline integrates dynamic DPP subset selection with adaptive embeddings to achieve complementary chunk selection.}
    \label{fig:pipeline}
\end{figure*}
% \subsection{The~\MyModel~Approach}
% \label{subsec:mymodel}
% \subsection{Overview}
An effective RAG system is ultimately governed not by how many candidates are retrieved, but by how much useful, distinct, and complementary information those chunks collectively convey. Under a fixed token budget, the generator can only take advantage of what is present in the candidate set. Thus, redundancy directly reduces the informational density of the augmentation, while homogeneity constrains unique evidence. %Existing RAG systems construct these contexts by independently measuring candidates according to user query relevance, implicitly treating retrieval as a point-wise scoring task rather than a set construction problem. However, this formulation fails to account for interactions among candidates and provides no mechanism to control redundancy or enforce complementarity. Thus, the retrieved context is often dominated by semantically overlapping chunks, even after reranking. 
To address this structural limitation, \textit{we frame the objective of retrieval as the task of constructing a subset whose elements are not only relevant to the query but also mutually diverse.} Guided by the formulation, we propose~\MyModel~, which jointly optimizes informational density, i.e., the relevance between query-candidates, and complementarity uniqueness, i.e., non-redundancy among candidates, within the available context window. %operating in two main phases: offline preparation and online inference. 
The overview is illustrated in Fig.~\ref{fig:pipeline}. 
%Now we elaborate on each design.

\subsection{DPP-based Subset Selection}
\label{subsec:dpp_subset_selection}
DPPs are probabilistic models for selecting diverse subsets, originating in statistical physics~\cite{macchi1975coincidence} and random matrix theory~\cite{hough2006determinantal}. 
%In machine learning, DPPs have been applied to diversity-promoting tasks such as recommendation systems~\cite{kulesza2012determinantal}, subset sampling~\cite{kulesza2012learningdeterminantalpointprocesses}, and in-context learning (ICL)~\cite{ye2023compositionalexemplarsincontextlearning}. 
Formally, given a ground set $\mathcal{Y} = \{1,2, \cdots, N \}$ with $N$ chunks, a DPP defines a probability distribution over all subsets $Y \subseteq \mathcal{Y}$ given by
\begin{equation}
    \small
    P(Y) = \frac{\det(\mathbf{L}_Y)}{\det(\mathbf{L} + \mathbf{I})},
    \label{eq:dpp_prob}
\end{equation}
where $\mathbf{L} \in \mathbb{R}^{N \times N}$ is a positive semi-definite (PSD) kernel matrix capturing chunk similarities and ensuring non-negative probabilities, $\mathbf{L}_Y$ denotes the submatrix of $\mathbf{L}$ indexed by $Y$, and $\mathbf{I}$ is the identity matrix. The PSD property of $\mathbf{L}$ ensures  $\det(\mathbf{L}_Y) \geq 0$ for any $\mathbf{Y} \subseteq \mathcal{Y}$. The normalization constant satisfies $\det(\mathbf{L} + \mathbf{I}) = \sum_{Y' \subseteq \mathcal{Y}} \det(\mathbf{L}_{Y'})$ with $\det(\mathbf{L}_\emptyset) = 1$ by convention, guaranteeing that $P(\cdot)$ defines a valid probability distribution over all subsets. Mathematically, $\mathbf{L}$ can be factorized as $\mathbf{L} = V^\top V$, where component $v_i \in \mathbb{R}^d$ of $V$ is the representation of $i$-th chunk. Under this view, the submatrix $\det(\mathbf{L}_Y)$ measures the squared volume spanned by the representation of chunks in $Y$. Therefore, subsets with a larger determinant, also the larger probability in Eq.~(\ref{eq:dpp_prob}), are those whose feature representations are more linearly independent, i.e., more diverse and closer to being orthogonal~\cite{hough2006determinantal}.

Motivated by the characteristics of DPPs, we propose a DPP-based subset selection mechanism to replace the standard top-$k$ selection in the conventional retrieval stage in the RAG system, thereby promoting informational density and uniqueness of candidates.   
% with a DPP-based mechanism that promotes diverse and complementary subsets from the candidate pool $\mathcal{D}^c$. This addresses the limitations of relevance-based selection by maximizing the determinant of the kernel matrix, which geometrically favors subsets with low redundancy and high information coverage. 

% This mechanism is especially beneficial for multi-hop reasoning, where the determinant maximization favors subsets that chain complementary information across chunks, supporting complex inference.
While the classic DPPs utilize a fixed kernel $\mathbf{L}$, which is difficult to adapt to a general RAG system, we first vary the kernel $\mathbf{L}$ dynamically on $\mathcal{D}^c$ by our P-Adapter, where $\mathcal{D}^c$ refers to $N$ retrieved chunks from knowledge base $\mathcal{D}= \{\mathbf{D}_1, \cdots, \mathbf{D}_M \}$. Given the initial representation $v_i$ of $i$-th chunk $\mathbf{D}_i$ that is mapped by an embedding model $\mathbb{E}(\cdot)$, we apply the P-Adapter $\boldsymbol{\phi}(\cdot)$ to obtain adapted embeddings $\hat{v}_i$. Thereby, the kernel $\mathbf{L}$ is updated as $\mathbf{L} = \hat{V}^\top \hat{V}$. % where the inner product captures pairwise similarities. 
For scale invariance, we %optionally 
normalize the embeddings to unit norm, making $\mathbf{L}$ equivalent to a cosine similarity matrix. Subsequently, the effective kernel updated by \MyModel~can be written as $\mathbf{\Gamma} = \mathbf{Q} \mathbf{L} \mathbf{Q}$. Herein, if a reranker is used, we incorporate query-chunk relevance by forming the diagonal quality matrix $\mathbf{Q} = \operatorname{diag}(\mathbf{q}_1, \cdots, \mathbf{q}_N)$, where $\mathbf{q}_i = \sqrt{\mathbf{s}_i}$ and the $\mathbf{s}_i$ are the positive reranker scores. Otherwise, we set $\mathbf{Q} = \mathbf{I}$. Finally, we employ the Maximum a Posteriori (MAP) to conduct the subset selection as:
% $\mathcal{D}^s$, namely:
\begin{equation}
    \mathcal{D}^s = \underset{Y \subseteq \mathcal{Y}, |Y|=k}{\arg\max}  \det(\mathbf{\Gamma}_{\mathcal{D}^s}).
\end{equation}
Herein, $\mathcal{D}^s$ is the selection subset of size $k$ from $\mathcal{D}^c$ that maximizes $P(Y)$ among all possible subsets $Y$ of size $k$. Notably, the probability $P(Y)$ is proportional to the determinant of the sub-kernel matrix $\mathbf{\Gamma}_{\mathcal{D}^s}$. This selection ensures that the chosen chunks are not only relevant but also non-redundant and synergistic, as the adapted embeddings $\hat{V}$ that are refined by the P-Adapter encode inter-chunk relations, thereby providing the LLM with optimized contexts for generation. Since exact MAP is NP-hard, we employ a fast greedy MAP inference algorithm  from~\cite{chen2018fast}. More details are incorporated in \underline{Appendix~\ref{ap:dpp_infer}}.

% Under this context, the effective kernel is $\mathbf{K} = \mathbf{Q} \mathbf{L} \mathbf{Q}$, balancing relevance (via $\mathbf{Q}$) and diversity (via $\mathbf{L}$). 
% , relying solely on the diversity encoded in $L'$.

% Given the initial embeddings $V \in \mathbb{R}^{N \times d}$ for $C$ (extracted during retrieval), we first apply the adapter $\phi$ (enabled here) to obtain adjusted embeddings $\phi(V')$. The kernel is then defined as $L'_{ij} = \phi(v'_i)^\top \phi(v'_j)$, where the inner product captures pairwise similarities. For scale invariance, we optionally normalize the embeddings to unit norm, making $L'$ equivalent to a cosine similarity matrix.

% \textcolor{red}{Thus we obtain candidate pool $\mathcal{D}^s$, which consists of $k$-largest submatrices of determinants.} 

% This approximates the max-det problem with submodular guarantees, incrementally maintaining the Cholesky decomposition of $K_S$ to evaluate marginal gains in $\mathcal{O}(k^2 N)$ time. 

% For numerical stability, we use $\epsilon = 10^{-6}$ and log-scaling in argmax. 

\subsection{P-Adapter}
\label{subsec:adapter}

Typically, retrieval in RAG systems is formulated as a point-wise scoring problem: an embedding model $\mathbb{E}(\cdot)$ maps the query $\mathbf{q}$ and each chunk $\mathbf{D}_i$ into a unified semantic space, the relevance is measured independently via a similarity function such as inner product or cosine. Candidates are then selected by ranking these scores and taking the top-$k$ chunks. This process treats each chunk in isolation, implicitly assuming that relevance between $\mathbf{q}$ and $\mathbf{D}_i$ is sufficient for constructing informative evidence. Consequently, inter-chunk interactions are neither modeled nor discoverable. %reranker 

To this end, we present a lightweight P-Adapter that enables standard embeddings with the capacity to encode inter-chunk complementarity without retraining the underlying encoder~\cite{houlsby2019parameter}. Specifically, we implement a feed-forward network with a bottleneck architecture as,
\begin{equation}
\small
    \boldsymbol{\phi}(v) = W_2 (GELU(LN(W_1 v + b_1))) + b_2, 
\end{equation}
where $v \in \mathbb{R}^d$ is the representation of a chunk (cf.~\ref{subsec:dpp_subset_selection}), $W_1 \in \mathbb{R}^{4d \times d}$ and $W_2 \in \mathbb{R}^{d \times 4d}$. It is worth mentioning that the P-Adapter is disabled during initial retrieval, ensuring that the relevance ranking remains intact. When constructing the DPP kernel, it is enabled. This targeted deployment allows the determinant maximization to operate over representations that encode inter-chunk interactions, biasing subset selection toward dense yet complementary contexts without perturbing the retrieval stage.

Consequently, training P-Adapter is performed on tuples $(\mathbf{q}, Y, \mathcal{N})$, where $\mathbf{q}$ is a query, $Y$ is the ground-truth positive complementary subset, $\mathcal{N}$ is a negative subset. %(e.g., redundant chunks sampled from candidates). %To construct $\mathcal{N}$, we remove $Y$ from the knowledge base and retrieve the top-($10-|Y|$) chunks using the frozen base encoder (BGE-M3~\cite{chen2024bge} in this work).  

\subsection{Diverse Margin Loss}
\label{subsec:dml}
While the DPP framework provides a principled mechanism for constructing diverse subsets, it alone does not specify how the embedding space should be shaped to reflect task-specific notions of complementarity. In particular, off-the-shelf embeddings are optimized for point-wise relevance, leaving no learning signal to distinguish truly complementary evidence chains from clusters of redundant yet individually relevant chunks. To bridge this gap, we introduce the Diverse Margin Loss (DML), a set-level objective that directly aligns the P-Adapter with the downstream subset construction goal. Formally, our DML can be expressed as:
% margin-like objective that penalizes cases where the determinant of negative subsets exceeds that of positives:
\begin{equation}
\small
    \mathcal{L}_{\text{DML}} = \big[ \max_{Y' \subseteq \mathcal{N}, |Y'|=k} \big( \det(\mathbf{L}_{Y'}) - \det(\mathbf{L}_Y) \big) \big]^+,
    \label{eq:dml}
\end{equation}
where $[\cdot]^+ = \text{ReLU}(\cdot)$ ensures non-negativity, penalizing only violations by the strongest negative subset while allowing natural similarities in positives. This targets bottlenecks from similar negatives, benchmarking against $\det(\mathbf{L}_Y)$ to encourage higher determinants for positive subsets.
% $Y$ is the ground-truth positive complementary subset, $\mathcal{N}$ is the set of negatives (e.g., redundant chunks sampled from candidates), 
% $Y'$ is a $k$-size subset from $\mathcal{N}$. %and $L$ is the positive semi-definite DPP kernel matrix.
Nevertheless, the objective of Eq.~(\ref{eq:dml}) is non-differentiable due to the $\max$ and $\text{ReLU}$. We derive a smooth approximation to enable gradient-based optimization as follows.

Initially, we approximate the $\max(\cdot)$ of Eq.~(\ref{eq:dml}) with LogSumExp (LSE), a convex, differentiable upper bound, as: %(see \underline{Appendix~\ref{ap:lse_approx}} for derivation):
\begin{equation}
\small
    \max_i x_i \approx \frac{1}{\gamma} \log \big( \sum_i \exp(\gamma x_i) \big),
\end{equation}
where $\gamma > 0$ controls sharpness (approaching true max as $\gamma \to \infty$, smoother for $\gamma \approx 1$). Applied to $\max_{Y'} \det(\mathbf{L}_{Y'})$:
\begin{equation}
\small
\max_{Y'} \det(\mathbf{L}_{Y'}) \approx \frac{1}{\gamma} \log \Big( \sum_{Y'} \exp \big( \gamma \det(\mathbf{L}_{Y'}) \big) \Big).
\end{equation}
Substituting yields:
\begin{equation}
\small
    \mathcal{L}_{\text{DML}} \approx \Big[ \frac{1}{\gamma} \log \Big( \sum_{Y'} \exp \big( \gamma \det(\mathbf{L}_{Y'}) \big) \Big) - \det(\mathbf{L}_{Y}) \Big]^+,
\end{equation}
converting the max to a differentiable sum-exp form.

Subsequently, we integrate the subtraction internally to enhance compactness and numerical stability by rewriting:
\begin{equation}
\small
    \begin{split}
    &\frac{1}{\gamma} \log \Big( \sum_{Y'} \exp \big( \gamma \det(\mathbf{L}_{Y'}) \big) \Big) - \det(\mathbf{L}_{Y}) \\
    &= \frac{1}{\gamma} \Big[ \log \Big( \sum_{Y'} \exp \big( \gamma \det(\mathbf{L}_{Y'}) \big) \Big) - \gamma \det(\mathbf{L}_{Y}) \Big].
    \end{split}
\end{equation}
Using $\log(a) - c = \log(a / \exp(c))$:
\begin{equation}
\small
    \begin{split}
    &\log \Big( \sum_{Y'} \exp \big( \gamma \det(\mathbf{L}_{Y'}) \big) \Big) - \gamma \det(\mathbf{L}_{Y}) \\
    &= \log \bigg( \sum_{Y'} \exp \Big( \gamma \big(\det(\mathbf{L}_{Y'}) - \det(\mathbf{L}_{Y}) \big) \Big) \bigg).
    \end{split}
\end{equation}
This yields:
\begin{equation}
\footnotesize
    \mathcal{L}_{\text{DML}} \approx \Bigg[ \frac{1}{\gamma} \log \bigg( \sum_{Y'} \exp \Big( \gamma \big(\det(\mathbf{L}_{Y'}) - \det(\mathbf{L}_{Y}) \big) \Big) \bigg) \Bigg]^+,
\end{equation}
moving the subtraction inside the exp for a cohesive structure.

Whereafter, we replace ReLU with softplus for full differentiability:
\begin{equation}
\small
[x]^+ \approx \log(1 + \exp(x)),
\end{equation}
a smooth upper bound with non-zero gradients ($\text{sigmoid}(x) \in (0,1)$). Substituting $x = \frac{1}{\gamma} \log\left( \sum \exp(\gamma (\det(\mathbf{L}_{Y'}) - \det(\mathbf{L}_Y))) \right)$:
\begin{equation}
    \footnotesize
    \begin{aligned}
    &\mathcal{L}_{\text{DML}} \approx \log \bigg( 1 + \\
    & \exp \Big( \frac{1}{\gamma} \log \big( \sum_{Y'} \exp( \gamma (\det(\mathbf{L}_{Y'}) - \det(\mathbf{L}_{Y})) ) \big) \Big) \bigg).
\end{aligned}
\end{equation}

% \small
%     \begin{split}
%     &\mathcal{L}_{\text{DML}} \approx \log \\
%     &\left( 1 + \exp\left( \frac{1}{\gamma} \log \left( \sum_{Y'} \exp( \gamma (\det(L_{Y'}) - \det(L_{Y})) ) \right) \right) \right).
%     \end{split}
% \end{equation}
Let $\tau$ denotes $\sum \exp(\gamma (\det(\mathbf{L}_{Y'}) - \det(\mathbf{L}_Y)))$, we can simplify $\exp ( \frac{1}{\gamma} \log(\tau) ) = \tau^{1/\gamma}$, namely:
\begin{equation}
    \small
    \mathcal{L}_{\text{DML}} \approx \log \Big( 1 + \Big[ \sum_{Y'} \exp \big( \gamma (\det(\mathbf{L}_{Y'}) - \det(\mathbf{L}_{Y})) \big) \Big]^{1/\gamma} \Big).
\end{equation}
Considering the balance of smoothness and accuracy, we typically set $\gamma=1$. Due to the space limitation, more details about the approximation proof can be found in \underline{Appendix~\ref{ap:proofs}}.
% Considering the balance of smoothness and accuracy, set $\gamma=1$ we can obtain:
% \begin{equation}
%     \small
%     \label{eq:dml_approx}
%     \mathcal{L}_{\text{DML}} = \log \left( 1 + \sum_{Y' \subseteq \mathcal{N}, |Y'|=k} \exp( \det(L_{Y'}) - \det(L_{Y}) ) \right).
% \end{equation}
% \begin{equation}
% L_{\text{DML}} \approx \log \left( 1 + \sum_{Y'} \exp( \det(L_{Y'}) - \det(L_{Y}) ) \right).
% \end{equation}
% The rationale is that softplus provides soft penalization with small gradients always.
% The final differentiable form is
% \begin{equation}
% L_{\text{DML}} = \log \left( 1 + \sum_{Y' \subseteq \mathcal{N}, |Y'|=k} \exp\left( \gamma (\det(L_{Y'}) - \det(L_{Y})) \right) \right)
% \end{equation}
% (with $\gamma \approx 1$).

\section{Experiments}
\label{experiment}
\begin{table*}[t]
\caption{Comparison on MultiHop-RAG benchmark, where the best result is highlighted in \textbf{bold}, and the second best with \underline{underline}.}
\label{tab:performance}
\setlength{\tabcolsep}{2pt}
\resizebox{\textwidth}{!}{
\begin{tabular}{@{}l*{12}{c}@{}}
\toprule
  & \multicolumn{6}{c}{\textbf{Without Reranker}}
  & \multicolumn{6}{c}{\textbf{With BAAI/bge-reranker-v2-m3}} \\
\cmidrule(lr){2-7} \cmidrule(lr){8-13}
  & NDCG@10 & Recall@10 & Hits@10 & NDCG@4 & Recall@4 & Hits@4
  & NDCG@10 & Recall@10 & Hits@10 & NDCG@4 & Recall@4 & Hits@4 \\
\midrule
BGE-Large\textsuperscript{\dag} (Standard RAG) & 0.4053 & 0.5923 & 0.6447 & 0.3354 & 0.4046 & 0.4860
  & 0.5909 & 0.7356 & 0.7587 & 0.5479 & 0.6232 & 0.6737 \\
\qquad + DPP Base, no adapter & 0.2133 & 0.4171 & 0.4860 & 0.1303 & 0.1870 & 0.2492
  & 0.5156 & 0.6196 & 0.6670 & 0.4735 & 0.5047 & 0.5765 \\
\qquad + \MyModel & 0.4359 & 0.6917 & 0.7274 & 0.3816 & 0.5416 & 0.6123
  & 0.6070 & 0.7210 & 0.7453 & \underline{0.5815} & \underline{0.6525} & 0.6983 \\
\midrule
BGE-m3\textsuperscript{\ddag} (Standard RAG) & 0.4259 & 0.6050 & 0.6827 & 0.3545 & 0.4142 & 0.5162
  & 0.5905 & 0.7344 & 0.7698 & 0.5454 & 0.6159 & 0.6860 \\
\qquad + DPP Base, no adapter & 0.2290 & 0.4215 & 0.5307 & 0.1439 & 0.1944 & 0.2782
  & 0.5154 & 0.6153 & 0.6939 & 0.4689 & 0.4939 & 0.5978 \\
\qquad + \MyModel & \underline{0.4631} & \underline{0.7182} & \underline{0.7620} & 0.3963 & 0.5387 & \underline{0.6358}
  & \underline{0.6089} & 0.7480 & 0.7765 & 0.5719 & 0.6502 & \underline{0.7240} \\
\midrule
Qwen3-0.6B\textsuperscript{\S} (Standard RAG) & 0.4147 & 0.6177 & 0.6749 & 0.3419 & 0.4186 & 0.5017
  & 0.5786 & 0.7339 & 0.7609 & 0.5344 & 0.6157 & 0.6682 \\
\qquad + DPP Base, no adapter & 0.2329 & 0.4056 & 0.4994 & 0.1645 & 0.2184 & 0.3017
  & 0.4896 & 0.5831 & 0.6480 & 0.4498 & 0.4751 & 0.5598 \\
\qquad + \MyModel & 0.4467 & 0.6784 & 0.7184 & \underline{0.4004} & \underline{0.5556} & 0.6313
  & 0.5884 & 0.7173 & 0.7419 & 0.5598 & 0.6413 & 0.6983 \\
\midrule
Qwen3-4B\textsuperscript{\P} (Standard RAG) & 0.4588 & 0.6669 & 0.7274 & 0.3799 & 0.4534 & 0.5441
  & 0.6036 & \underline{0.7732} & \underline{0.8045} & 0.5490 & 0.6285 & 0.7006 \\
\qquad + DPP Base, no adapter & 0.2265 & 0.4052 & 0.5117 & 0.1497 & 0.2047 & 0.2972
  & 0.4963 & 0.5881 & 0.6626 & 0.4570 & 0.4825 & 0.5788 \\
\qquad + \MyModel & \textbf{0.4895} & \textbf{0.7453} & \textbf{0.7866} & \textbf{0.4339} & \textbf{0.5942} & \textbf{0.6793}
  & \textbf{0.6326} & \textbf{0.7855} & \textbf{0.8123} & \textbf{0.5980} & \textbf{0.6907} & \textbf{0.7497} \\
\bottomrule
\end{tabular}
}
{\footnotesize
\textsuperscript{\dag}BGE-Large $\to$ BAAI/bge-large-en-v1.5 \quad
\textsuperscript{\ddag}BGE-m3 $\to$ BAAI/bge-m3 \quad
\textsuperscript{\S}Qwen3-0.6B $\to$ Qwen/Qwen3-Embedding-0.6B \quad
\textsuperscript{\P}Qwen3-4B $\to$ Qwen/Qwen3-Embedding-4B
}
\vspace{-4mm}
\end{table*}

% \subsection{Setup}
% \label{subsec:setup}

\noindent{\textbf{Benchmark and Metrics.}}
In this work, we evaluate~\MyModel~as a plug-in enhancement module on the MultiHop-RAG benchmark~\cite{tang2024multihoprag}, a challenging dataset for multi-hop question answering consisting of 2,556 queries derived from news articles, covering inference, temporal, and comparison reasoning across 2-hop to 4-hop settings. This benchmark is well-suited for evaluating inter-chunk complementarity, as correct evidence must be jointly retrieved across chained contexts. We exclude 301 null queries with empty evidence, resulting in 2,255 valid instances. The dataset is split into training, validation, and test sets in a 5:1:4 ratio using stratified sampling over hop counts. For retrieval performance, we report standard IR metrics: Recall at K (Recall@K), Normalized Discounted Cumulative Gain at K (NDCG@K), and Hit Rate at K (Hits@K).%Retrieval performance is evaluated using standard IR metrics, including Recall@K, NDCG@K, and Hits@K.
% We evaluate~\MyModel~as an enhancement module on the MultiHop-RAG benchmark~\cite{tang2024multihoprag}, a challenging dataset designed for multi-hop question answering, comprising 2,556 queries derived from news articles with complex multi-hop reasoning types, including inference, temporal, and comparison questions across 2-hop, 3-hop, and 4-hop structures. This benchmark is particularly suitable for assessing inter-chunk complementarity, as effective subsets must cover chained information across chunks. We exclude 301 null queries (with empty evidence lists and "Insufficient information" answers), resulting in 2,255 effective queries. For training and evaluation, we split the dataset using stratified sampling based on hop counts (2-hop, 3-hop, and 4-hop) to maintain proportional distributions, resulting in a 5:1:4 ratio for train, validation, and test sets, respectively. For retrieval performance, we report standard IR metrics: Recall at K (Recall@K), Normalized Discounted Cumulative Gain at K (NDCG@K), and Hit Rate at K (Hits@K). 
% These capture ranking quality, recall, and hit rates.
% Mean Reciprocal Rank at K (MRR@K), Mean Average Precision at K (MAP@K),

\noindent{\textbf{Implementations.}} 
We evaluate our \MyModel~using four representative embedding backbones: BAAI/bge-large-en-v1.5~\cite{bge_embedding}, BAAI/bge-m3~\cite{chen2024bge}, Qwen/Qwen3-Embedding-0.6B, and Qwen/Qwen3-Embedding-4B~\cite{qwen3embedding}. All methods use BAAI/bge-reranker-v2-m3~\cite{chen2024bge} for reranking. The candidate pool size is fixed to $N=20$, with subset selection at $k=10$ or $k=4$. Experiments are conducted on a single RTX 5090 GPU, using a learning rate of $1\mathrm{e}{-4}$ for 20 epochs. Batch sizes are set to 8 for all models except Qwen3-Embedding-4B (batch size 4). Training takes approximately 0.3 hours for the smaller models and 1.5 hours for Qwen3-Embedding-4B. Embedding dimensions are 1024 for all models except Qwen3-Embedding-4B (2560)~\footnote{Our code is available at \url{https://anonymous.4open.science/r/ScalDPP-8E92}.}.

% We employ four embedding models: BAAI/bge-large-en-v1.5~\cite{bge_embedding}, BAAI/bge-m3~\cite{chen2024bge}, Qwen/Qwen3-Embedding-0.6B~\cite{qwen3embedding}, and Qwen/Qwen3-Embedding-4B~\cite{qwen3embedding}. For reranking, we uniformly use BAAI/bge-reranker-v2-m3~\cite{chen2024bge}, a lightweight multilingual reranker model with 568M parameters. The candidate pool size is set to $N=20$, with subset selection at $k=10$ or $k=4$. All experiments were conducted on an RTX 5090. The learning rate was set to 0.0001 across all models, with training running for 20 epochs. For the BAAI/bge-large-en-v1.5, BAAI/bge-m3, and Qwen/Qwen3-Embedding-0.6B models, the batch size was set to 8, while for Qwen/Qwen3-Embedding-4B, it was set to 4. Training times were approximately 0.3 hours for BAAI/bge-large-en-v1.5, BAAI/bge-m3, and Qwen/Qwen3-Embedding-0.6B, and about 1.5 hours for Qwen/Qwen3-Embedding-4B. The embedding dimensions are 1024 for BAAI/bge-large-en-v1.5, BAAI/bge-m3, and Qwen/Qwen3-Embedding-0.6B, and 2560 for Qwen/Qwen3-Embedding-4B.

% \subsection{Results}
% \label{subsec:results}
%Without reranker, it achieves average relative improvements of 7.7\% in NDCG@10 (e.g., 0.4588 to 0.4895 on Qwen3-4B, +6.7\%), 14.3\% in Recall@10 (e.g., 0.6669 to 0.7453 on Qwen3-4B, +11.8\%), and 9.8\% in Hits@10. Gains persist at $k=4$ (NDCG@4 +14.2\%, Recall@4 +31.9\%, Hits@4 +25.0\%), where determinant maximization favors orthogonal subsets, mitigating token dilution in constrained contexts, as in our motivation example (cf. Fig.~\ref{fig:motivation}). 
\noindent{\textbf{Main Results.}} 
Table~\ref{tab:performance} shows that \MyModel\ consistently outperforms standard RAG across all embedding backbones and evaluation metrics on MultiHop-RAG. In particular, we have the following findings: Without reranking, it yields consistent relative improvements, averaging +7.7\% in NDCG@10 (e.g., from 0.4588 to 0.4895 on Qwen3-4B), +14.3\% in Recall@10 (e.g., from 0.6669 to 0.7453), and +9.8\% in Hits@10. The advantages become more pronounced under stricter context budgets ($k=4$), with gains of +14.2\% in NDCG@4, +31.9\% in Recall@4, and +25.0\% in Hits@4, where determinant-based subset selection favors orthogonal evidence and mitigates token redundancy (cf. Fig.~\ref{fig:motivation}). 
When a reranker is applied, \MyModel\ maintains improvements, yielding an average gain of +3.1\% in NDCG@10 (e.g., 0.6036 → 0.6326 on Qwen3-4B), indicating that diversity-aware selection complements relevance-focused reranking via the fused quality matrix $\mathbf{Q}$. Gains also scale with backbone capacity: Qwen3-0.6B attains +7.7\% NDCG@10 without reranking, while Qwen3-4B obtains the strongest results (Hits@4 = 0.7497 with reranking). Overall, results highlight the value of explicitly modeling inter-chunk diversity and complementarity for multi-hop evidence aggregation. Due to the limited space, the end-to-end QA evaluation that shows the capacity of \MyModel~on downstream generations is provided in \underline{Appendix~\ref{ap:qa_analysis}}. The results echo our statement that modeling the interaction between candidates benefits the construction of ground evidence. 

% With reranker, gains average 3.1\% in NDCG@10 (e.g., 0.6036 to 0.6326 on Qwen3-4B, +4.8\%), amplified by $Q$ fusing relevance with diversity. Performance scales with model size: Qwen3-0.6B shows +7.7\% NDCG@10 without reranker, while Qwen3-4B peaks (Hits@4=0.7497 with reranker). These results validate inter-chunk modeling for multi-hop evidence chaining.

\noindent{\textbf{Ablation Study.}} To uncover the effectiveness of P-Adapter in \MyModel, Table~\ref{tab:performance} also presents the detailed ablation results. We have the following observations:
Removing the adapter ("DPP Base, no adapter") causes substantial drops: -53.7\% NDCG@10 (0.2265 vs. 0.4895), -45.6\% Recall@10, and -34.9\% Hits@10 on Qwen3-4B, worsening at $k=4$ (-65.5\% NDCG@4, -65.6\% Recall@4, -56.2\% Hits@4), highlighting the adapter’s role in injecting positive relations via DML and enabling complementarity. The reranker improves all variants, synergizing with~\MyModel~(+32.9\% NDCG@10 average; e.g., BGE-m3 Hits@4 +13.9\% from 0.6358 to 0.7240), while DPP Base shows larger relative gains (+124.0\% NDCG@10) from a lower baseline, emphasizing the adapter’s preconditioning for $\mathbf{Q}$ fusion. Even without reranking,~\MyModel~outperforms standard RAG (+7.7\% NDCG@10), demonstrating its efficacy even without any reranking refinement.

\begin{figure}[htbp]
\centering
\includegraphics[width=0.6\linewidth]{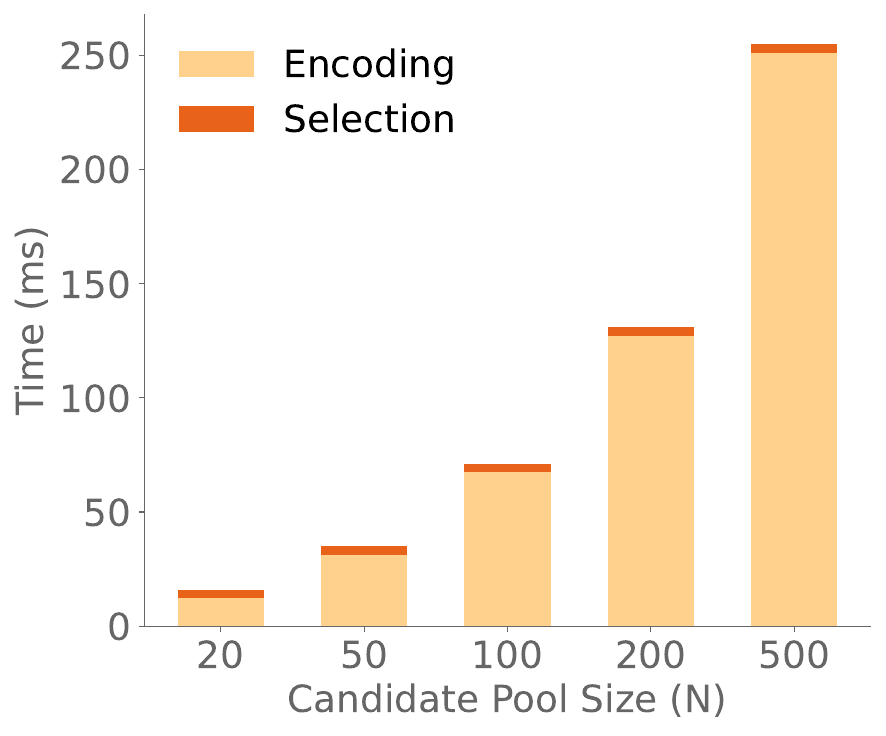}
\caption{Time consumption analysis.} %Breakdown of runtime components for \MyModel, showing the time spent on embedding computation and kernel matrix construction/inference, with fixed $k=10$ and varying candidate pool sizes $N$.}
\label{fig:efficiency_time_breakdown}
\vspace{-2mm}
\end{figure}

\noindent \textbf{Efficiency.}
\label{ap:efficiency}
The time consumption of \MyModel~primarily arises from two components: 1) mapping chunks to semantic space, and 2) dynamically constructing kernel matrix $\mathbf{\Gamma}$ and performing Fast Greedy MAP Inference. To evaluate the efficiency of \MyModel, we analyze its runtime under varying candidate pool sizes $N$. As illustrated in Fig.~\ref{fig:efficiency_time_breakdown}, the overall latency grows approximately linearly with $N$ and is dominated by the encoding stage. In contrast, the cost of selection remains consistently small across all settings, indicating that the additional set-level modeling introduced by \MyModel~is computationally lightweight and does not become a bottleneck.

% Ablating the adapter ("DPP Base, no adapter") yields average drops of -50.8\% in NDCG@10 (e.g., 0.2265 vs. 0.4895 on Qwen3-4B, -53.7\%), -41.7\% in Recall@10, and -32.2\% in Hits@10 without reranker, worsening at $k=4$ (-63.5\% NDCG@4, -63.9\% Recall@4, -56.0\% Hits@4). This underscores the adapter's role in injecting positive relations via DML training, overcoming PSD constraints for complementarity in multi-hop paths. Reranker boosts all variants, but synergizes with~\MyModel~(+32.9\% NDCG@10 average from no- to with-reranker; e.g., BGE-m3 Hits@4 +13.9\% from 0.6358 to 0.7240). DPP Base gains more (+124.0\% NDCG@10) from low baseline, highlighting adapter's preconditioning for $\mathbf{Q}$ fusion. Without reranker,~\MyModel~still exceeds standard RAG (+7.7\% NDCG@10), affirming efficacy in low-resource settings for factual downstream generation.

% \subsection{Analysis}
% \label{subsec:analysis}

\begin{figure}[htbp]
\centering
\subfloat[DML without reranker.]{
\includegraphics[width=0.4\linewidth]{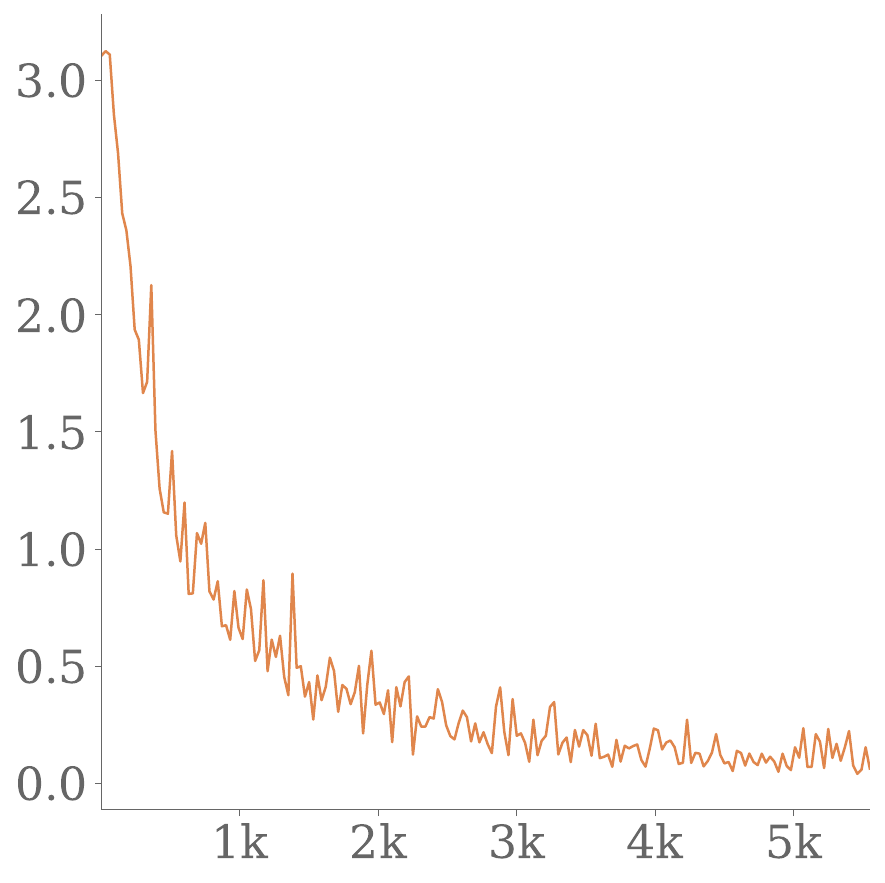}
\label{fig:dml_curve_no_reranker}
}
\subfloat[DML with reranker.]{
\includegraphics[width=0.4\linewidth]{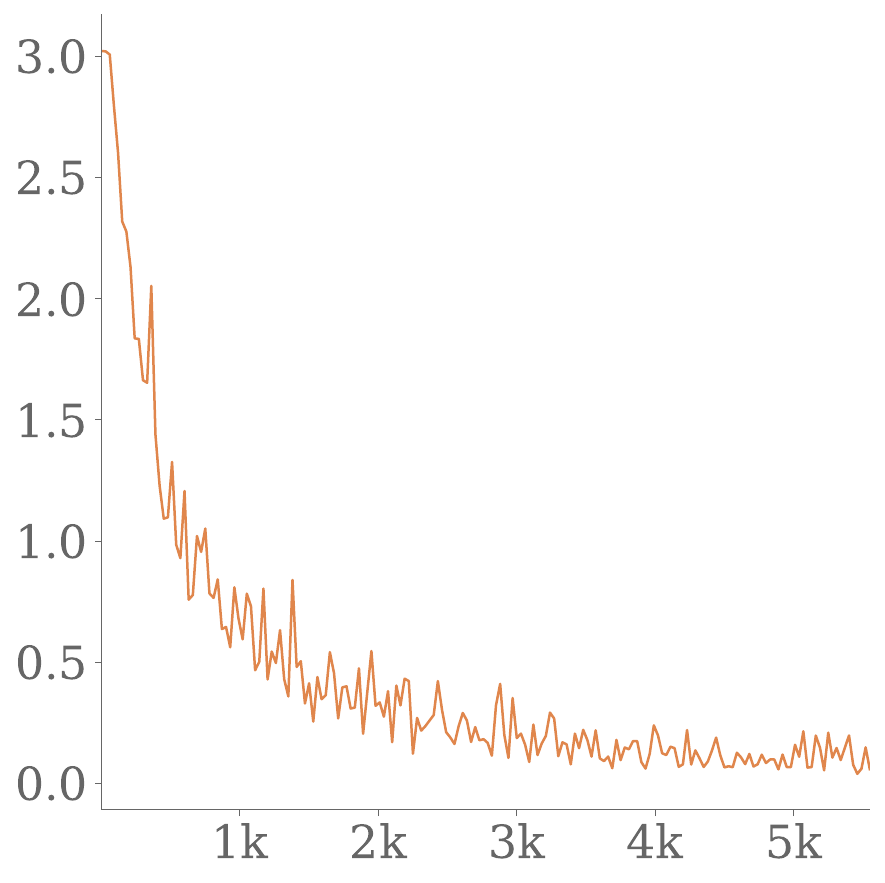}
\label{fig:dml_curve_with_reranker}
}

\subfloat[NLL without reranker.]{
\includegraphics[width=0.4\linewidth]{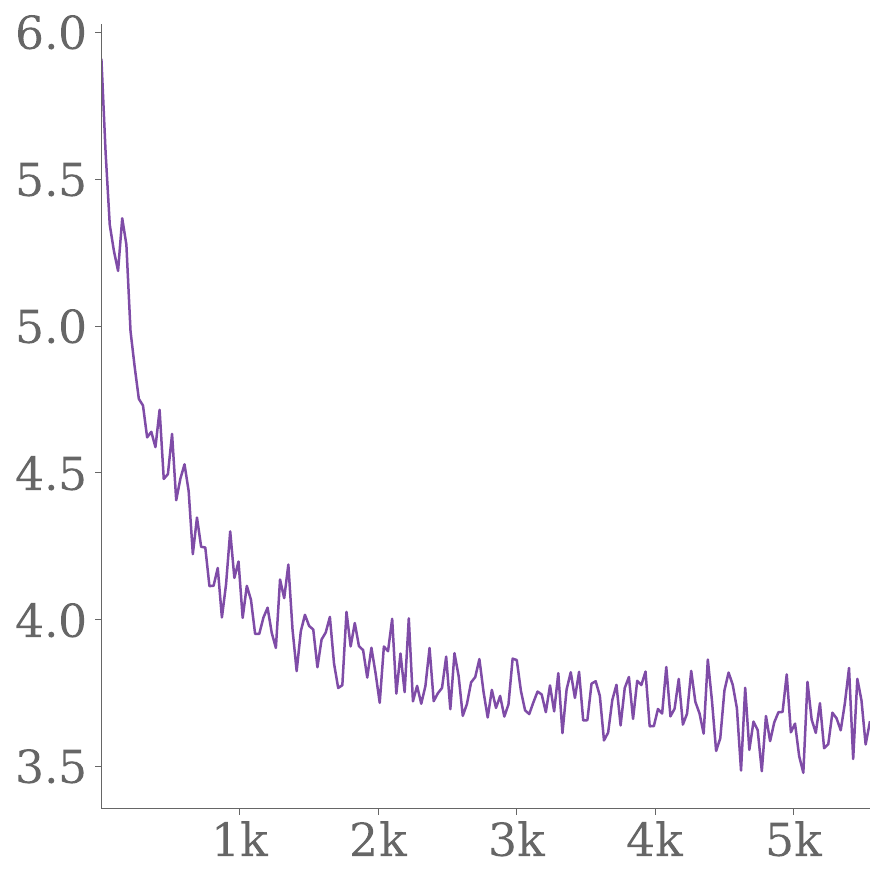}
\label{fig:NLL_curve_no_reranker}
}
\subfloat[NLL with reranker.]{
\includegraphics[width=0.4\linewidth]{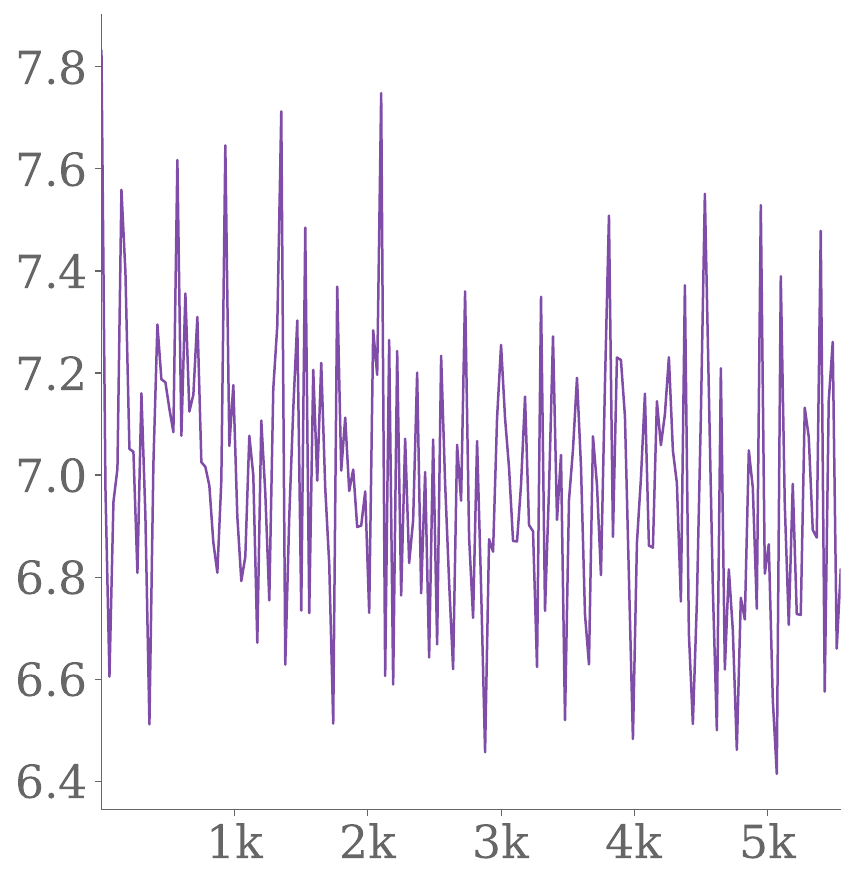}
\label{fig:NLL_curve_with_reranker}
}
\caption{Training curves for DML and NLL with/without reranker.}
\label{fig:training_curves}
\vspace{-4mm}
\end{figure}
% Logging step 20, total epoch 20

\noindent{\textbf{Training Curve Differences.}} 
To further understand the observed differences in training dynamics between DML and NLL, we analyze their behaviors through the lens of their mathematical formulations and optimization properties. DML's smooth approximation, given by
\begin{equation}
    \footnotesize
    \begin{split}
    & \mathcal{L}_{\text{DML}}(\theta) = \log \bigg(1 + \\
    & \sum_{Y' \subseteq \mathcal{N}, |Y'|=k} \exp\! \Big( \det \big( \mathbf{L}_{Y'}(\theta) \big) - \det \big(\mathbf{L}_Y(\theta) \big) \Big) \bigg),
    \end{split}
\end{equation}
where \(\theta\) denotes the trainable parameters of the adapter, incorporates a margin-based penalty via the relative differences between determinants of positive and negative subsets.\footnote{In practice, the sum over subsets may be approximated via sampling for large \(m = \binom{|\mathcal{N}|}{k}\) to ensure computational feasibility.} This design yields a near-convex loss landscape in the parameter space \(\theta\), as the approximation serves as a convex upper bound on the original non-differentiable objective (proven in Appendix~\ref{ap:convex_bound}, where convexity holds with respect to the determinant values). The gradient structure, which includes weighted contributions from negative subsets (via soft argmax-like weights, \(\frac{\partial \mathcal{L}_{\text{DML}}}{\partial \theta} \propto \sum_{Y'} w_{Y'} \cdot \frac{\partial}{\partial \theta} (\det(\mathbf{L}_{Y'}) - \det(\mathbf{L}_Y))\), with \(w_{Y'} = \frac{\exp(\det(\mathbf{L}_{Y'}) - \det(\mathbf{L}_Y))}{1 + \sum \exp(\cdot)}\)), provides stable and informative updates, preventing overshooting and ensuring minimal oscillations. This weighted summation acts as a form of attention over violators, focusing gradients on the most problematic negative subsets while maintaining bounded variance, even in high-dimensional \(\theta\) spaces (e.g., \(d=1024\) or 2560). Moreover, its scale-invariant nature, focusing on relative determinant differences, mitigates numerical sensitivity, making it robust to variations in embedding magnitudes or reranker scores. This results in rapid convergence, as seen in Figures~\ref{fig:dml_curve_no_reranker} and~\ref{fig:dml_curve_with_reranker}, even with reranker integration, where quality scores \(\mathbf{Q}\) (with \(\mathbf{q}_i = \sqrt{\mathbf{s}_i}\)) are robustly absorbed into the relative differences, preserving well-conditioned Hessian eigenvalues (consistent with the positive semi-definite Hessian of log-sum-exp~\cite{boyd2004convex}) and avoiding saddle points common in non-convex settings.

In contrast, NLL, defined as the negative log-likelihood with normalization,
\begin{equation}
\small
\mathcal{L}_{\text{NLL}}(\theta) = -\log\det(\mathbf{L}_Y(\theta)) + \log\det(\mathbf{L}(\theta) + \mathbf{I}),
\end{equation}
where \(\theta\) again represents the adapter parameters, exhibits a highly non-convex landscape due to the inherent non-linearity of the log-determinant function. The gradients involve opposing forces: the first term maximizes diversity in the positive subset, while the second provides global regularization via the normalization constant \(Z = \det(\mathbf{L} + \mathbf{I})\). This opposition leads to high gradient variance, as \(\nabla \mathcal{L}_{\text{NLL}} = -\nabla\log\det(\mathbf{L}_Y) + \nabla\log\det(\mathbf{L} + \mathbf{I})\), where the terms can counteract each other, especially in high-dimensional embeddings (e.g., 1024 or 2560 dimensions), resulting in net gradients with erratic directions. The Hessian of \(\log\det(\mathbf{M})\), given by \(\nabla^2 \log\det(\mathbf{M}) = -\mathbf{M}^{-1} \otimes \mathbf{M}^{-1}\) (Kronecker product)~\cite{boyd2004convex}, becomes ill-conditioned near degenerate matrices (condition number exploding as eigenvalues approach zero), amplifying small perturbations in \(\theta\) into large loss fluctuations. Additionally, scaling mismatches arise because \(\det(\mathbf{L}_Y)\) (for small \(k=2 \sim 4\)) and \(\det(\mathbf{L} + \mathbf{I})\) (for larger \(N=10\)) span vastly different magnitudes, leading to gradient explosions or vanishings when determinant values approach degeneracy. Without reranker, the simplicity of base embeddings allows eventual convergence, albeit with larger oscillations from occasional determinant instabilities, as the identity-dominant \(\mathbf{L} + \mathbf{I}\) provides some stabilization. With reranker, the introduced \(\mathbf{Q}\) matrix (with \(\mathbf{q}_i = \sqrt{\mathbf{s}_i}\)) exacerbates these issues by distorting eigenvalues in \(\mathbf{L} + \mathbf{I}\), causing persistent oscillations and convergence challenges, as evident in Figures~\ref{fig:NLL_curve_no_reranker} and~\ref{fig:NLL_curve_with_reranker}. These differences underscore DML's superior stability for multi-hop RAG tasks, where capturing complementary relations requires robust optimization over chained, high-variance subsets, and NLL's vulnerabilities highlight the need for relative, margin-based designs in such probabilistic subset selection problems.

\begin{table*}[t]
\caption{Performance of~\MyModel~by hop count on MultiHop-RAG using BAAI/bge-m3 under different losses.}%, where the best overall result is highlighted in \textbf{bold}
\label{tab:hop_performance}
\resizebox{\textwidth}{!}{
\begin{tabular}{@{}ll*{12}{c}@{}}
\toprule
 & & \multicolumn{6}{c}{\textbf{Without Reranker}}
  & \multicolumn{6}{c}{\textbf{With BAAI/bge-reranker-v2-m3}} \\
\cmidrule(lr){3-8} \cmidrule(lr){9-14}
 & & NDCG@10 & Recall@10 & Hits@10 & NDCG@4 & Recall@4 & Hits@4
  & NDCG@10 & Recall@10 & Hits@10 & NDCG@4 & Recall@4 & Hits@4 \\
\midrule
\multirow{4}{*}{Standard}
& 2-hop & 0.5000 & 0.6671 & 0.7220 & 0.4452 & 0.5199 & 0.6168 & 0.6301 & 0.7652 & 0.7850 & 0.5874 & 0.6542 & 0.7056 \\
& 3-hop & 0.3503 & 0.5352 & 0.6234 & 0.2708 & 0.3220 & 0.4091 & 0.5320 & 0.6926 & 0.7370 & 0.4843 & 0.5649 & 0.6429 \\
& 4-hop & 0.3729 & 0.5734 & 0.6918 & 0.2722 & 0.3082 & 0.4528 & 0.5971 & 0.7322 & 0.7925 & 0.5504 & 0.6116 & 0.7170 \\
& overall & 0.4259 & 0.6050 & 0.6827 & 0.3545 & 0.4142 & 0.5162 & 0.5905 & 0.7344 & 0.7698 & 0.5454 & 0.6159 & 0.6860 \\
\midrule
\multirow{4}{*}{DML}
& 2-hop & 0.4519 & 0.7044 & 0.7477 & 0.3822 & 0.5140 & 0.5981 & 0.6254 & 0.7500 & 0.7804 & 0.5869 & 0.6425 & 0.7150 \\
 & 3-hop & 0.4640 & 0.7267 & 0.7597 & 0.3989 & 0.5519 & 0.6494 & 0.5749 & 0.7430 & 0.7662 & 0.5365 & 0.6461 & 0.7208 \\
 & 4-hop & 0.4913 & 0.7385 & 0.8050 & 0.4296 & 0.5797 & 0.7107 & 0.6304 & 0.7521 & 0.7862 & 0.5998 & 0.6787 & 0.7547 \\
 & overall & \textbf{0.4631} & \textbf{0.7182} & \textbf{0.7620} & \textbf{0.3963} & \textbf{0.5387} & \textbf{0.6358} & \textbf{0.6089} & \textbf{0.7480} & \textbf{0.7765} & \textbf{0.5719} & \textbf{0.6502} & \textbf{0.7240} \\
\midrule
\multirow{4}{*}{NLL\textsuperscript{\dag}}
& 2-hop & 0.4295 & 0.6542 & 0.7150 & 0.3666 & 0.4871 & 0.5841 & 0.5992 & 0.6881 & 0.7196 & 0.5673 & 0.6016 & 0.6589 \\
 & 3-hop & 0.4666 & 0.7067 & 0.7500 & 0.4257 & 0.6001 & 0.6883 & 0.5908 & 0.7522 & 0.7695 & 0.5576 & 0.6656 & 0.7208 \\
 & 4-hop & 0.4449 & 0.6667 & 0.7233 & 0.3971 & 0.5514 & 0.6730 & 0.6175 & 0.7558 & 0.7799 & 0.5770 & 0.6488 & 0.7296 \\
 & overall & 0.4450 & 0.6745 & 0.7285 & 0.3924 & 0.5374 & \textbf{0.6358} & 0.5995 & 0.7222 & 0.7475 & 0.5657 & 0.6320 & 0.6927 \\
\bottomrule
\end{tabular}
}
{\footnotesize
\textsuperscript{\dag}NLL: Log-Determinant Loss, which maximizes \(\log \det(\mathbf{L}_Y)\) for positive subsets \(Y\).
% \begin{equation}
% \log \det(\mathbf{L}_Y) - \log \det(\mathbf{L} + I)
% \end{equation}
}
\vspace{-2mm}
\end{table*}

\noindent{\textbf{Loss Function Comparison.}} We now compare our devised DML against the standard Negative Log-Likelihood (NLL) loss regarding MultiHop-RAG performance. As also shown in Table~\ref{tab:hop_performance}, DML consistently outperforms NLL across metrics by incorporating margin-based penalties on negative subsets, enabling better capture of positive complementary relations beyond NLL's pure maximization of $\log \det(\mathbf{L}_Y)$. Without a reranker, DML improves overall NDCG@10 by 4.1\% over NLL (0.4631 vs. 0.4450), with larger gains in Recall@10 (+6.5\%, 0.7182 vs. 0.6745) and Hits@10 (+4.6\%, 0.7620 vs. 0.7285), particularly evident in 4-hop queries (+10.4\% NDCG@10, 0.4913 vs. 0.4449). With a reranker, DML yields a 1.6\% NDCG@10 boost (0.6089 vs. 0.5995), amplified in Hits@4 (+4.5\%, 0.7240 vs. 0.6927). Both surpass Standard baselines (DML +8.7\% NDCG@10 overall without reranker), but DML's focus on penalizing redundant negatives results in more informative subsets for multi-hop evidence chaining. As illustrated in Figure~\ref{fig:training_curves}, the training curves further highlight DML's advantages: DML converges quickly with minimal oscillations both without and with reranker, whereas NLL converges with larger fluctuations without reranker and exhibits persistent oscillations with reranker, making convergence challenging.
\begin{figure*}[htbp]
    \centering
    \includegraphics[width=\linewidth]{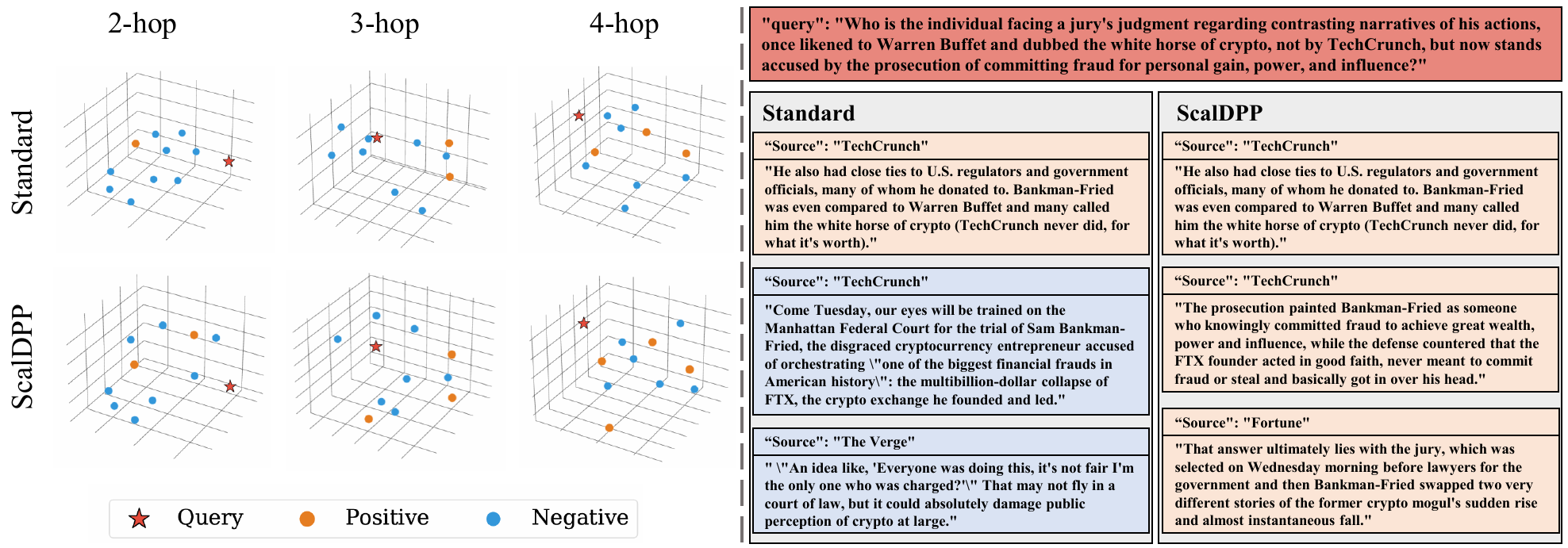}
    \caption{Case study on multi-hop queries. Left: t-SNE projections of chunk embeddings for 2-, 3-, and 4-hop cases; top row: Standard RAG, bottom row:~\MyModel. Right: Zoom-in on the 3-hop query showing the query text and the top-3 retrieved chunks. Standard RAG selects only one ground-truth chunk in its top-3, whereas~\MyModel~precisely recovers all three required positive evidence chunks.}
    \label{fig:case}
    \vspace{-3mm}
\end{figure*}

\noindent{\textbf{Performance by Hop Count.}} 
% As shown in Table~\ref{tab:hop_performance}, \MyModel~with DML achieves larger performance gains for higher-hop queries, where inter-chunk complementarity is critical for chained evidence. Without a reranker, improvements increase with hop count, and are further amplified under constrained contexts ($k=4$), mitigating token dilution. Trends persist with reranking, highlighting DPP’s effectiveness in promoting diverse, synergistic subsets for multi-hop reasoning.%
In our approach, the proposed DML serves as a set-level objective, directly aligning the P-Adapter with the downstream subset selection task. To validate this design, 
as shown in Table~\ref{tab:hop_performance}, \MyModel\ with DML exhibits increasingly larger performance gains as query complexity grows, particularly in higher-hop scenarios where inter-chunk complementarity is critical for chaining evidence. Without a reranker, DML yields a substantial relative improvement of 31.8\% in NDCG@10 for 4-hop queries (0.4913 vs. 0.3729 for Standard), compared to a slight degradation for 2-hop queries and strong gains for 3-hop queries, resulting in an average improvement of 8.7\% across all hop counts. These advantages become more pronounced under constrained context budgets ($k=4$), where DML achieves a 57.8\% gain in 4-hop NDCG@4, effectively mitigating token dilution. When a reranker is applied, the same trend persists, with DML delivering a 5.6\% NDCG@10 improvement on 4-hop queries and an average gain of 3.1\% overall, underscoring the effectiveness of DPP-based selection in promoting diverse and complementary evidence for multi-hop reasoning.
%As shown in Table~\ref{tab:hop_performance},~\MyModel~with DML exhibits increasing performance gains with query complexity, particularly in higher-hop scenarios where inter-chunk complementarity is essential for chained evidence paths. Without a reranker, DML achieves a 31.8\% relative improvement in NDCG@10 for 4-hop queries (0.4913 vs. 0.3729 for Standard), compared to a -9.6\% for 2-hop (0.4519 vs. 0.5000) and 32.5\% for 3-hop (0.4640 vs. 0.3503), yielding an average of 8.7\% across all hops. These gains are amplified at $k=4$ (e.g., 4-hop NDCG@4 +57.8\% from 0.2722 to 0.4296), mitigating token dilution in constrained contexts. With a reranker, trends persist with DML delivering 5.6\% NDCG@10 improvement in 4-hop (0.6304 vs. 0.5971), averaging 3.1\% overall, underscoring DPP's efficacy in promoting diverse, synergistic subsets for multi-hop reasoning.

%To qualitatively demonstrate the effectiveness of~\MyModel~in capturing inter-chunk complementarity and diversity, we present t-SNE~\cite{maaten2008visualizing} visualizations and determinant analysis on representative 2-, 3-, and 4-hop queries from MultiHop-RAG (Fig.~\ref{fig:case}).

% , also substantiating our statement in practice.

% a fixed subset size of $k=10$ with varying candidate pool sizes $N$ (e.g., 20, 50, 100, 200, 500). All timings are measured in milliseconds on a single RTX 5090 GPU.

\noindent{\textbf{Case Study.}} 
To qualitatively illustrate \MyModel’s ability to capture inter-chunk diversity and complementarity, we present t-SNE~\cite{maaten2008visualizing} visualizations and determinant analyses on representative 2-, 3-, and 4-hop queries from MultiHop-RAG (cf. Fig.~\ref{fig:case}). All t-SNE projections are computed on the original BGE-M3 embeddings, so the query point and ground-truth positive chunks occupy identical positions across methods. The visual differences thus arise solely from the distribution of surrounding scatter points (negative/irrelevant chunks) and the final selected subsets. We have the following findings:

% All t-SNE projections are computed on the original BGE-M3 embeddings, so the query point and ground-truth positive chunks occupy identical positions across methods. The visual differences therefore arise solely from the distribution of surrounding scatter points (negative/irrelevant chunks) and the final selected subsets.

(1) In the top row (Standard RAG), selected chunks tightly cluster around the query, reflecting a strong bias toward proximity. This often results in incomplete ground-truth coverage, as redundant but semantically similar chunks crowd out distant yet complementary evidence. (2) In the bottom row (our~\MyModel), the selected subsets exhibit markedly greater dispersion while still encompassing all ground-truth positive chunks in every case (2-, 3-, and 4-hop). This demonstrates that the adapter successfully reshapes the embedding space to favor orthogonal, complementary directions over mere query similarity. (3) The right panel zooms into a challenging 3-hop query (full text shown). Standard RAG recovers only one of the three required positive chunks in its top-3, filling the rest with nearby but redundant or tangential articles. In contrast, our~\MyModel~precisely retrieves all three complementary evidence chunks, forming a complete chained path despite some being farther from the query in the original embedding space.

Table~\ref{tab:det_analysis} provides quantitative evidence: in the original BGE-M3 space, the margin \(\det(\mathbf{L}_Y) - \max\det(\mathbf{L}_{Y'})\) is consistently negative, meaning redundant negative subsets span larger subspace volumes than the ground-truth positives. After adapter transformation, the margin becomes strongly positive across all hop counts, with negative determinants collapsing near zero while the positive subset volume is dramatically enlarged. This geometric reinterpretation directly explains the superior recovery of complementary evidence.

% These visualizations and metrics illustrate how~\MyModel~overcomes the proximity bias of standard dense retrieval, enabling robust multi-source evidence chaining essential for complex multi-hop reasoning and delivering richer, non-redundant context to the LLM.

\begin{table}[ht]
\centering
\caption{Determinant Analysis of Subspace Volumes in Original BGE-M3 ($e_i$) and BGE-M3 with Adapter ($\phi(e_i)$) Embedding Spaces on MultiHop-RAG.}
\label{tab:det_analysis}
\resizebox{0.85\linewidth}{!}{
\begin{tabular}{@{}llccc@{}}
\toprule
 & & 2-hop & 3-hop & 4-hop \\
\midrule
\multirow{4}{*}{\textbf{$e_i$}}
& $\det(\mathbf{L}_Y)$                  & 0.7396 & 0.2664 & 0.1457 \\
& $\max\det(\mathbf{L}_Y')$             & 0.8451 & 0.3430 & 0.2560 \\
& $mean \det(\mathbf{L}_Y')$            & 0.6649 & 0.2458 & 0.1940 \\
& $std \det(\mathbf{L}_Y')$             & 0.1261 & 0.0505 & 0.0423 \\
& $margin$                     & -0.1055 & -0.0766 & -0.1103 \\
\midrule
\multirow{4}{*}{\textbf{$\phi(e_i)$}}
& $\det(\mathbf{L}_Y)$                  & 0.9982 & 0.0519  & 0.0154  \\
& $\max\det(\mathbf{L}_Y')$             & 0.3342 & 0.0004  & 0.0000  \\
& $mean \det(\mathbf{L}_Y')$            & 0.1402 & 0.0001 & 0.0000 \\
& $std \det(\mathbf{L}_Y')$             & 0.1117 & 0.0001 & 0.0000 \\
& $margin$                     & \textbf{0.6640} & \textbf{0.0515} & \textbf{0.0154} \\
\bottomrule
\end{tabular}
}
{\footnotesize 
$margin = \det(\mathbf{L}_Y) - \max\det(\mathbf{L}_Y')$ 
}
\vspace{-4mm}
\end{table}

%\section{}
\section{Related Work}
\label{related_work}

% Retrieval-Augmented Generation (RAG) enhances LLMs by dynamically incorporating external knowledge during generation, effectively alleviating limitations of parametric knowledge in knowledge-intensive tasks and improving factual accuracy while reducing hallucinations~\cite{lewis2020retrieval, guu2020realm}. 
Modern Retrieval-Augmented Generation (RAG) systems typically adopt hybrid retrieval pipelines that combine sparse lexical matching methods (e.g., BM25~\cite{robertson2009probabilistic}) with dense semantic retrievers (e.g., Contriever~\cite{izacard2021contriever}) and high-quality embedding models (e.g., BGE~\cite{chen2024bge}, Qwen3 Embedding~\cite{qwen3embedding}) to construct high-recall candidate sets, followed by rerankers trained to capture query–chunk relevance~\cite{li2023making, chen2024bge}. However, these approaches predominantly model relevance between the query and individual chunks in isolation, neglecting inter-chunk interactions. This query-centric formulation is particularly limiting in multi-hop reasoning scenarios (e.g., MultiHop-RAG~\cite{tang2024multihoprag}).
Fortunately, Determinantal Point Processes (DPPs), originally studied in statistical physics and random matrix theory~\cite{macchi1975coincidence, hough2006determinantal}, provide a probabilistic framework for modeling repulsive interactions and have been widely used in machine learning to promote diversity~\cite{kulesza2012determinantal}, including applications in recommendation~\cite{wilhelm2018practical}, summarization~\cite{cho2019multi}, diverse generation~\cite{elfeki2019gdpp}, and in-context learning~\cite{ye2023compositionalexemplarsincontextlearning}. DPPs select subsets by modeling negative dependencies among items via kernel determinants, with learning typically based on likelihood maximization and inference performed using k-DPPs or greedy approximations~\cite{kulesza2011kdpp, chen2018fast}. However, standard DPPs are computationally expensive since the need to pre-train the full kernel matrix, which limits scalability to large knowledge bases. Their positive semi-definite constraint also restricts interactions to repulsion, preventing the modeling of complementary relationships among chunks.

\section{Conclusion}
\label{conclusion}
We present \MyModel, a novel framework that integrates Determinantal Point Processes (DPPs) into RAG to capture inter-chunk diversity and complementarity. By combining a scalable dynamic kernel, a parameter-efficient P-Adapter, and a set-level Diverse Margin Loss (DML), \MyModel\ significantly addresses standard DPPs’ scalability and correlation limitations while improving multi-hop subset selection. Experiments on MultiHop-RAG show consistent gains across embeddings, hop counts, and reranking setups. Our work highlights the importance of inter-chunk interactions in RAG and provides a plug-and-play approach for constructing diverse, informative contexts.

% \section*{Acknowledgements}

% \textbf{Do not} include acknowledgements in the initial version of the paper

% If a paper is accepted, the final camera-ready version can (and usually should)
% include acknowledgements.  Such acknowledgements should be placed at the end of
% the section, in an unnumbered section that does not count towards the paper
% page limit. Typically, this will include thanks to reviewers who gave useful
% comments, to colleagues who contributed to the ideas, and to funding agencies
% and corporate sponsors that provided financial support.

\section*{Impact Statement}

This paper presents work whose goal is to advance the field of Machine
Learning. There are many potential societal consequences of our work, none
which we feel must be specifically highlighted here.

% % In the unusual situation where you want a paper to appear in the
% % references without citing it in the main text, use \nocite

\bibliography{refs}
\bibliographystyle{icml2026}

%%%%%%%%%%%%%%%%%%%%%%%%%%%%%%%%%%%%%%%%%%%%%%%%%%%%%%%%%%%%%%%%%%%%%%%%%%%%%%%
%%%%%%%%%%%%%%%%%%%%%%%%%%%%%%%%%%%%%%%%%%%%%%%%%%%%%%%%%%%%%%%%%%%%%%%%%%%%%%%
% APPENDIX
%%%%%%%%%%%%%%%%%%%%%%%%%%%%%%%%%%%%%%%%%%%%%%%%%%%%%%%%%%%%%%%%%%%%%%%%%%%%%%%
%%%%%%%%%%%%%%%%%%%%%%%%%%%%%%%%%%%%%%%%%%%%%%%%%%%%%%%%%%%%%%%%%%%%%%%%%%%%%%%
\newpage
\appendix
\onecolumn % 开启则附录以单栏形式呈现，关闭则以双栏形式呈现
% \section{Appendix}
\section*{APPENDIX}

\section{Availability}
\label{ap:availability}
The source code for~\MyModel~is publicly available at \url{https://anonymous.4open.science/r/ScalDPP-8E92} under the MIT License.

\section{Derivation of Fast Greedy MAP Inference}
\label{ap:dpp_infer}

The fast greedy MAP inference algorithm is in algorithm~\ref{alg:greedy_map}, achieving $\mathcal{O}(k^2 N)$ time complexity for selecting a subset of cardinality $k$ from $N$ candidates~\cite{chen2018fast}. In this paper's PyTorch implementation, engineering optimizations were adopted while strictly preserving algorithmic equivalence and correctness. These include replacing per-candidate loops with vectorized computation, incrementally expanding the lower-triangular Cholesky factor $C$ for stable triangular solves. These modifications enhance numerical stability and leverage GPU acceleration for large-scale candidate sets.

\begin{algorithm}[htbp]
\caption{Fast Greedy MAP Inference for DPP}
\label{alg:greedy_map}
\small
\begin{algorithmic}[1]
\STATE \textbf{Input:} Kernel matrix $\mathbf{\Gamma} \in \mathbb{R}^{N \times N}$, target subset size $k$
\STATE \textbf{Initialize:} $d_i^2 = \mathbf{\Gamma}_{ii}$, $\mathbf{c}_i = []$, $\forall i \in \mathcal{Y}$, $j = \arg\max_{i \in \mathcal{Y}} \log(d_i^2 + \epsilon)$, $Y_g = \{j\}$

\WHILE{$|Y_g| < k$ \textbf{and} $\max_{i \in \mathcal{Y}} d_i^2 > \epsilon$}
    \FOR{$i \in \mathcal{Y} \setminus Y_g$}
        \STATE $e_i = \mathbf{\Gamma}_{j i} - \mathbf{c}_j^\top \mathbf{c}_i / d_j^2$
        \STATE $\mathbf{c}_i = [\mathbf{c}_i \;\; e_i]$
        \STATE $d_i^2 = \max(d_i^2 - e_i^2, 0)$
    \ENDFOR
    \STATE $j = \arg\max_{i \in \mathcal{Y} \setminus Y_g} \log(d_i^2 + \epsilon)$
    \STATE $Y_g = Y_g \cup \{j\}$
\ENDWHILE

\STATE \textbf{return} $Y_g$
\end{algorithmic}
\end{algorithm}

\section{Proofs}
\label{ap:proofs}

\subsection{LSE Approximation of Maximum Function}
\label{ap:lse_approx}

The LogSumExp (LSE) function provides a smooth, differentiable approximation to the maximum operator, which is widely used in convex optimization and machine learning to handle non-differentiable objectives. Consider a set of values \(x_1, x_2, \dots, x_m \in \mathbb{R}\). The scaled LSE is defined as
\begin{equation}
\text{LSE}_\gamma(x) = \frac{1}{\gamma} \log \left( \sum_{i=1}^m \exp(\gamma x_i) \right),
\end{equation}
where \(\gamma > 0\) is a scaling parameter that controls the sharpness of the approximation. As \(\gamma \to \infty\), \(\text{LSE}_\gamma(x)\) approaches \(\max_i x_i\), while smaller \(\gamma\) (e.g., \(\gamma \approx 1\)) yields a smoother upper bound.

This approximation is discussed in~\cite{boyd2004convex}, where the unscaled log-sum-exp function \(f(x) = \log \left( \sum_{i=1}^m \exp(x_i) \right)\) is shown to be convex and to satisfy
\begin{equation}
\max_i x_i \leq f(x) \leq \max_i x_i + \log m.
\end{equation}
The scaled version extends this by tightening the bound as \(\gamma\) increases.

To prove that \(\text{LSE}_\gamma(x)\) is an upper bound on \(\max_i x_i\), let \(x^* = \max_i x_i\). We start by noting that for each \(i\), \(x_i \leq x^*\), so \(\exp(\gamma x_i) \leq \exp(\gamma x^*)\). Therefore,
\begin{equation}
\sum_{i=1}^m \exp(\gamma x_i) \leq \sum_{i=1}^m \exp(\gamma x^*) = m \exp(\gamma x^*).
\end{equation}
Taking the natural logarithm on both sides:
\begin{equation}
\log \left( \sum_{i=1}^m \exp(\gamma x_i) \right) \leq \log \left( m \exp(\gamma x^*) \right) = \log m + \gamma x^*.
\end{equation}
Dividing by \(\gamma > 0\):
\begin{equation}
\text{LSE}_\gamma(x) = \frac{1}{\gamma} \log \left( \sum_{i=1}^m \exp(\gamma x_i) \right) \leq x^* + \frac{1}{\gamma} \log m = \max_i x_i + \frac{1}{\gamma} \log m.
\end{equation}
This establishes the upper bound, with the error term \(\frac{1}{\gamma} \log m\) approaching 0 as \(\gamma \to \infty\).

For the lower bound, observe that the sum includes at least the maximum term:
\begin{equation}
\sum_{i=1}^m \exp(\gamma x_i) \geq \exp(\gamma x^*),
\end{equation}
since all terms are positive and the sum is at least as large as its largest term. Taking the logarithm:
\begin{equation}
\log \left( \sum_{i=1}^m \exp(\gamma x_i) \right) \geq \log \left( \exp(\gamma x^*) \right) = \gamma x^*.
\end{equation}
Dividing by \(\gamma\):
\begin{equation}
\text{LSE}_\gamma(x) \geq x^* = \max_i x_i.
\end{equation}
Thus, \(\max_i x_i \leq \text{LSE}_\gamma(x) \leq \max_i x_i + \frac{1}{\gamma} \log m\), confirming that \(\text{LSE}_\gamma(x)\) is a convex upper bound on the max, with the approximation tightening as \(\gamma\) increases or \(m\) decreases.

\subsection{Proof that the Approximation is a Convex Upper Bound}
\label{ap:convex_bound}

In this appendix, we prove that the approximated Diverse Margin Loss (DML) is a convex upper bound on the original non-differentiable objective. Recall the original loss:
\begin{equation}
\mathcal{L}_{\text{DML}} = \left[ \max_{Y' \subseteq \mathcal{N}, |Y'|=k} \det(\mathbf{L}_{Y'}) - \det(\mathbf{L}_Y) \right]^+,
\end{equation}
where \([x]^+ = \max(0, x)\).

The approximated form is:
\begin{equation}
\tilde{\mathcal{L}}_{\text{DML}} = \frac{1}{\gamma} \log \left( 1 + \sum_{Y' \subseteq \mathcal{N}, |Y'|=k} \exp\left( \gamma (\det(\mathbf{L}_{Y'}) - \det(\mathbf{L}_{Y})) \right) \right).
\end{equation}
For simplicity, we consider \(\gamma = 1\):
\begin{equation}
\tilde{\mathcal{L}}_{\text{DML}} = \log \left( 1 + \sum_{Y' \subseteq \mathcal{N}, |Y'|=k} \exp( \det(\mathbf{L}_{Y'}) - \det(\mathbf{L}_{Y}) ) \right),
\end{equation}
though the proof extends to general \(\gamma > 0\) by scaling the arguments appropriately (e.g., replacing exponents with \(\gamma\) multiples and adjusting the outer factor). Specifically, for general \(\gamma\), the bounds and convexity hold analogously because the scaling preserves the inequalities and convexity properties, as \(\gamma\) acts as a positive multiplier in the exponents and a divisor outside the log.

To treat convexity, we view the loss as a function of the determinants. Let \(s_p = \det(\mathbf{L}_Y)\) be the positive score, and let \(s_i = \det(\mathbf{L}_{Y'_i})\) for \(i = 1, \dots, m\) where \(m = \binom{|\mathcal{N}|}{k}\) is the number of negative subsets. The original is \([ \max_i s_i - s_p ]^+\), and the approximation is 
\begin{equation}
\tilde{\mathcal{L}}(s_1, \dots, s_m, s_p) = \log \left( 1 + \sum_{i=1}^m \exp( s_i - s_p ) \right).
\end{equation}
We prove that \(\tilde{\mathcal{L}} \geq \mathcal{L}\) (upper bound) and that \(\tilde{\mathcal{L}}\) is convex in \((s_1, \dots, s_m, s_p)\).

\subsubsection{Step 1: Prove the Upper Bound (\(\tilde{\mathcal{L}} \geq \mathcal{L}\))}
1. Denote \(z_i = s_i - s_p\) for each \(i\), so \(z = (z_1, \dots, z_m)\). Then the original loss is \([ \max_i z_i ]^+\), and the approximation is
   \begin{equation}
   \tilde{\mathcal{L}}(z) = \log \left( 1 + \sum_{i=1}^m \exp(z_i) \right).
   \end{equation}
   This substitution is valid because subtracting \(s_p\) from each \(s_i\) shifts all determinants by a constant, preserving relative differences.

2. The sum \(\sum_{i=1}^m \exp(z_i) \geq \exp( \max_i z_i )\), since the sum is at least the largest term (all terms positive):
   \begin{equation}
   \sum_{i=1}^m \exp(z_i) \geq \exp( \max_i z_i ),
   \end{equation}
   because for the index \(j\) where \(z_j = \max_i z_i\), the sum \(\geq \exp(z_j)\), and the remaining \(m-1\) terms are each \(\geq 0\) (as exponentials are always positive). Equality holds if all other \(z_k < z_j\) and their exponentials are negligible, or if there's only one term.

3. Adding 1 to both sides:
   \begin{equation}
   1 + \sum_{i=1}^m \exp(z_i) \geq 1 + \exp( \max_i z_i ).
   \end{equation}
   This preserves the inequality since 1 is positive and added equally.

4. Since \(\log\) is monotonically increasing:
   \begin{equation}
   \log \left( 1 + \sum_{i=1}^m \exp(z_i) \right) \geq \log \left( 1 + \exp( \max_i z_i ) \right).
   \end{equation}
   Note that both arguments to log are greater than 1, ensuring positivity.

5. The right-hand side is the softplus function: \(\log(1 + \exp( \max_i z_i )) = \text{softplus}( \max_i z_i )\).

6. Now, prove that \(\text{softplus}(w) \geq [w]^+\) for any \(w \in \mathbb{R}\).
\begin{itemize}
   \item \(w \geq 0\): We can rewrite this as:
     \begin{equation}
     \log(1 + \exp(w)) = \log(\exp(w) (\exp(-w) + 1)) = \log(\exp(w)) + \log(1 + \exp(-w)) = w + \log(1 + \exp(-w)).
     \end{equation}
     Since \(\exp(-w) > 0\) for all finite \(w\), it follows that \(1 + \exp(-w) > 1\), and thus \(\log(1 + \exp(-w)) > \log(1) = 0\). Therefore, \(\text{softplus}(w) = w + \text{positive number} > w = [w]^+\). The inequality is strict unless \(\exp(-w) \to 0\) (i.e., \(w \to \infty\)), where it approaches equality asymptotically.
   \item \(w < 0\): Since \(\exp(w) > 0\) (exponential is always positive), \(1 + \exp(w) > 1\), so \(\log(1 + \exp(w)) > \log(1) = 0\). Meanwhile, \([w]^+ = 0\) because \(w < 0\). Thus, \(\text{softplus}(w) > 0 = [w]^+\). Again, the inequality is strict, approaching equality as \(w \to -\infty\), where \(\exp(w) \to 0\) and \(\text{softplus}(w) \to 0\).
\end{itemize}

7. Setting \(w = \max_i z_i\), we have \(\text{softplus}( \max_i z_i ) \geq [ \max_i z_i ]^+\). This follows directly from the case analysis in step 6, applied to this specific \(w\).

8. Combining steps 4 and 7:
   \begin{equation}
   \tilde{\mathcal{L}}(z) \geq \text{softplus}( \max_i z_i ) \geq [ \max_i z_i ]^+ = \mathcal{L}.
   \end{equation}
   The chain of inequalities holds because each part is greater than or equal to the next, establishing the overall upper bound.
This establishes the upper bound. Note that this bound is tight in certain limits: for example, if one \(z_i\) dominates (much larger than others), the sum approximates \(\exp(\max z_i)\), and \(\tilde{\mathcal{L}} \approx \text{softplus}(\max z_i) \approx \max z_i\) when \(\max z_i \gg 0\); when \(\max z_i \to -\infty\), both approach 0. Additionally, for small \(m\), the bound is closer, but for large \(m\), the sum may inflate the value, providing a looser but still valid upper bound.
\subsubsection{Step 2: Prove Convexity of \(\tilde{\mathcal{L}}\)}
As shown in~\cite{boyd2004convex}, the log-sum-exp function \(f(w) = \log \left( \sum_{j=1}^{n} \exp(w_j) \right)\) is convex in \(w \in \mathbb{R}^n\). This is established through its Hessian being positive semidefinite, as detailed in the reference.
Our \(\tilde{\mathcal{L}}(z) = \log \left( 1 + \sum_{i=1}^m \exp(z_i) \right)\) can be expressed as log-sum-exp on an extended vector: let \(w = (0, z_1, \dots, z_m) \in \mathbb{R}^{m+1}\), then
\begin{equation}
\tilde{\mathcal{L}}(z) = f(w) = \log \left( \exp(0) + \sum_{i=1}^m \exp(z_i) \right).
\end{equation}
The mapping from \(z\) to \(w\) is affine (specifically, it embeds \(z\) into a higher-dimensional space with a fixed constant component: the first entry is always 0, independent of \(z\), while the rest are linear copies of \(z_i\)). Composition of a convex function with an affine mapping preserves convexity: if \(f\) is convex and \(A(z) = Bz + c\) where \(B\) is a matrix (here, \(B\) is a selection/embedding matrix) and \(c = (0, 0, \dots, 0)\) except for the first component, then \(f(A(z))\) is convex. To see why, for any \(\theta \in [0,1]\) and points \(z_1, z_2\),
\begin{equation}
A(\theta z_1 + (1-\theta) z_2) = \theta A(z_1) + (1-\theta) A(z_2),
\end{equation}
by linearity of affine maps, and then
\begin{equation}
f(A(\theta z_1 + (1-\theta) z_2)) = f(\theta A(z_1) + (1-\theta) A(z_2)) \leq \theta f(A(z_1)) + (1-\theta) f(A(z_2)),
\end{equation}
by the convexity of \(f\). Thus, \(\tilde{\mathcal{L}}(z)\) is convex in \(z\).
Now, recall that \(z_i = s_i - s_p\), so \(\tilde{\mathcal{L}}\) is a function of \((s_1, \dots, s_m, s_p)\). The mapping \((s_1, \dots, s_m, s_p) \mapsto (s_1 - s_p, \dots, s_m - s_p)\) is also affine: it can be represented as a linear transformation where each \(z_i = 1 \cdot s_i + (-1) \cdot s_p + 0 \cdot \text{others}\), forming a matrix with 1's on the diagonal for \(s_1\) to \(s_m\) and -1's in the \(s_p\) column. Composing the convex \(\tilde{\mathcal{L}}(z)\) with this affine map again preserves convexity, as per the same reasoning above. Therefore, \(\tilde{\mathcal{L}}(s_1, \dots, s_m, s_p)\) is convex in its arguments.

\section{End-to-End Question Answering Evaluation}
\label{ap:qa_analysis}

To evaluate the capacity of~\MyModel~on downstream generation, we conduct an end-to-end question answering (QA) task on the MultiHop-RAG benchmark. Using the retrieved contexts from the selected subsets, we prompt the DeepSeek-V3.2~\cite{deepseekai2025deepseekv32} (with thinking mode, limiting output tokens to 4096) to generate answers.
%for the multi-hop queries, 
Notably, the QA prompt follows the original work~\cite{tang2024multihoprag}. %MultiHop-RAG benchmark~\cite{tang2024multihoprag} for testing. 
We report standard QA metrics: Exact Match (EM@K), F1 score (F1@K), and Accuracy (Acc@K), where K represents the context budget (top-$10$ or top-$4$ chunks). These metrics quantify how well the optimized contexts reduce hallucinations and improve factual accuracy in the generated responses. Experiments are conducted with the same set of main results. %embedding backbones and reranking setups as in the main retrieval evaluation, ensuring direct comparability. 
Our \MyModel~consistently outperforms against baselines, especially without any reranker refinement. An interesting phenomenon is that the variation RAG+DPP Base w/o P-Adapter achieves better performance when the context budget is set to 10, supporting our claim that the DPP-based methods take advantage of diversity to generate, and further, under the limited context budget, redundant chunks will crowd out the informative context. 
\vspace{-4mm}
\begin{table*}[htbp]
\centering
\caption{End-to-End QA Performance Comparison on the MultiHop-RAG Benchmark using BAAI/bge-m3.} %The best result is highlighted in \textbf{bold}, and the second best with \underline{underline}.}
\label{tab:qa_performance}
\setlength{\tabcolsep}{2pt}
\resizebox{\textwidth}{!}{
\begin{tabular}{@{}ll*{12}{c}@{}}
\toprule
 & & \multicolumn{6}{c}{\textbf{Without Reranker}}
  & \multicolumn{6}{c}{\textbf{With BAAI/bge-reranker-v2-m3}} \\
\cmidrule(lr){3-8} \cmidrule(lr){9-14}
 & & EM@10 & F1@10 & Acc@10 & EM@4 & F1@4 & Acc@4
  & EM@10 & F1@10 & Acc@10 & EM@4 & F1@4 & Acc@4 \\
\midrule
\multirow{4}{*}{Standard RAG}
& 2-hop & 0.3855 & 0.3889 & 0.3995 & 0.2850 & 0.2882 & 0.2897
  & 0.4322 & 0.4357 & 0.4393 & 0.3738 & 0.3769 & 0.3762 \\
& 3-hop & 0.4383 & 0.4405 & 0.4383 & 0.3604 & 0.3611 & 0.3636
  & 0.4383 & 0.4405 & 0.4416 & 0.3701 & 0.3723 & 0.3701 \\
& 4-hop & 0.6981 & 0.7044 & 0.6981 & 0.6101 & 0.6164 & 0.6101
  & 0.7170 & 0.7261 & 0.7170 & 0.6415 & 0.6506 & 0.6415 \\
& overall & \underline{0.4592} & \underline{0.4627} & \underline{0.4659} & \underline{0.3687} & \underline{0.3716} & \underline{0.3721}
  & 0.4849 & 0.4889 & 0.4894 & 0.4201 & 0.4240 & 0.4212 \\
\midrule
\multirow{4}{*}{+ DPP Base, w/o P-Adapter}
& 2-hop & 0.3154 & 0.3170 & 0.3201 & 0.2150 & 0.2150 & 0.2150
  & 0.4509 & 0.4549 & 0.4673 & 0.3832 & 0.3866 & 0.3879 \\
& 3-hop & 0.4805 & 0.4805 & 0.4805 & 0.3831 & 0.3853 & 0.3831
  & 0.4740 & 0.4762 & 0.4740 & 0.3831 & 0.3874 & 0.3864 \\
& 4-hop & 0.7421 & 0.7449 & 0.7421 & 0.5786 & 0.5786 & 0.5786
  & 0.7170 & 0.7261 & 0.7170 & 0.7044 & 0.7135 & 0.7044 \\
& overall & 0.4480 & 0.4493 & 0.4503 & 0.3374 & 0.3382 & 0.3374
  & \textbf{0.5061} & \textbf{0.5104} & \textbf{0.5140} & \underline{0.4402} & \underline{0.4450} & \underline{0.4436} \\
\midrule
\multirow{4}{*}{+ \MyModel}
& 2-hop & 0.4065 & 0.4081 & 0.4136 & 0.2897 & 0.2928 & 0.2967
  & 0.4252 & 0.4275 & 0.4369 & 0.3762 & 0.3777 & 0.3832 \\
& 3-hop & 0.4740 & 0.4775 & 0.4805 & 0.4026 & 0.4069 & 0.4026
  & 0.4838 & 0.4838 & 0.4838 & 0.4123 & 0.4152 & 0.4156 \\
& 4-hop & 0.7233 & 0.7233 & 0.7233 & 0.6352 & 0.6415 & 0.6352
  & 0.7044 & 0.7166 & 0.7044 & 0.6918 & 0.7009 & 0.6918 \\
& overall & \textbf{0.4860} & \textbf{0.4880} & \textbf{0.4916} & \textbf{0.3899} & \textbf{0.3940} & \textbf{0.3933}
  & \underline{0.4950} & \underline{0.4982} & \underline{0.5006} & \textbf{0.4447} & \textbf{0.4480} & \textbf{0.4492} \\
\bottomrule
\end{tabular}
}
\end{table*}

\end{document}